%% file: 00-main.tex
\documentclass[runningheads]{llncs}
\usepackage[dvipsnames]{xcolor}
\usepackage{graphicx}
\usepackage[skip=5pt]{caption}

\usepackage{tikz}
\usepackage{booktabs}
\usepackage{comment}
\usepackage{amsmath,amssymb} 
\usepackage{multirow}
\usepackage{xspace}
\usepackage{bbm}

\usepackage[numbers,sort,compress]{natbib}

\usepackage[accsupp]{axessibility}  

\usepackage[pagebackref,breaklinks,colorlinks]{hyperref}
\usepackage{float}
\floatstyle{plaintop}
\restylefloat{table}

\usepackage[capitalize]{cleveref}
\crefname{section}{Sec.}{Secs.}
\Crefname{section}{Section}{Sections}
\Crefname{table}{Table}{Tables}
\crefname{table}{Tab.}{Tabs.}

\newcommand{\spotP}{P}
\newcommand{\spotMnew}{M$^{*}$}
\newcommand{\spotDnew}{D$^{*}$}
\newcommand{\spotExemplarKnown}{E}
\newcommand{\spotExemplarNovel}{N}

\def\sepappendix{0}

\begin{document}
\pagestyle{headings}
\mainmatter
\def\ECCVSubNumber{1028}  
\newcommand*\samethanks[1][\value{footnote}]{\footnotemark[#1]}
\title{Automatic dense annotation of\\large-vocabulary sign language videos}

\titlerunning{Automatic dense annotation}

\author{Liliane Momeni\inst{1}\thanks{Equal contribution} \and
Hannah Bull\inst{2}\samethanks \and
K R Prajwal\inst{1}\samethanks \and \\
Samuel Albanie\inst{3}\and
Gül Varol\inst{4}\and
Andrew Zisserman\inst{1}}
\authorrunning{L. Momeni et al.}

\institute{Visual Geometry Group, University of Oxford, UK \and
LISN, Univ Paris-Saclay, CNRS, France \and
Department of Engineering, University of Cambridge, UK \and
LIGM, École des Ponts, Univ Gustave Eiffel, CNRS, France \\
\email{\{liliane,prajwal,albanie,gul,az\}@robots.ox.ac.uk; hannah.bull@lisn.upsaclay.fr}
\url{https://www.robots.ox.ac.uk/~vgg/research/bsldensify/} 
}

\maketitle

\input{./01-abstract.tex}
\input{./02-intro.tex}
\input{./03-related.tex}

\input{./04-method.tex}
\input{./05-experiments.tex}

\input{./06-conclusion.tex}

\input{./08-acknowledgements.tex}

\bibliographystyle{splncs04}
\bibliography{refs}

\newpage
{\noindent \large \bf {APPENDIX}}\\
\input{07-appendix.tex}

\end{document}

%% file: 01-abstract.tex
\begin{abstract}

Recently, sign language researchers have turned to sign language interpreted TV broadcasts, comprising (i) a video of continuous signing and (ii) subtitles corresponding to the audio content, as a readily available and large-scale source of training data.
One key challenge in the usability of such data is the lack of sign annotations. Previous work exploiting such weakly-aligned data only found \textit{sparse} correspondences between keywords in the subtitle and individual signs. In this work, we propose a simple, scalable framework to \textit{vastly} increase the \textit{density} of automatic annotations. Our contributions are the following: (1)~we significantly improve previous annotation methods by making use of synonyms and subtitle-signing alignment; (2)~we show the value of pseudo-labelling from a sign recognition model as a way of sign spotting; (3)~we propose a novel approach for increasing our annotations of \textit{known} and \textit{unknown} classes based on \textit{in-domain exemplars}; (4)~on the BOBSL BSL sign language corpus, we increase the number of confident automatic annotations from 670K to 5M. We make these annotations publicly available to support the sign language research community.

\keywords{Sign Language Recognition, Automatic Dataset Construction, Novel Class Discovery.}
\end{abstract}

%% file: 02-intro.tex
\begin{figure}[t]
    \centering
    \includegraphics[width=0.90\textwidth]{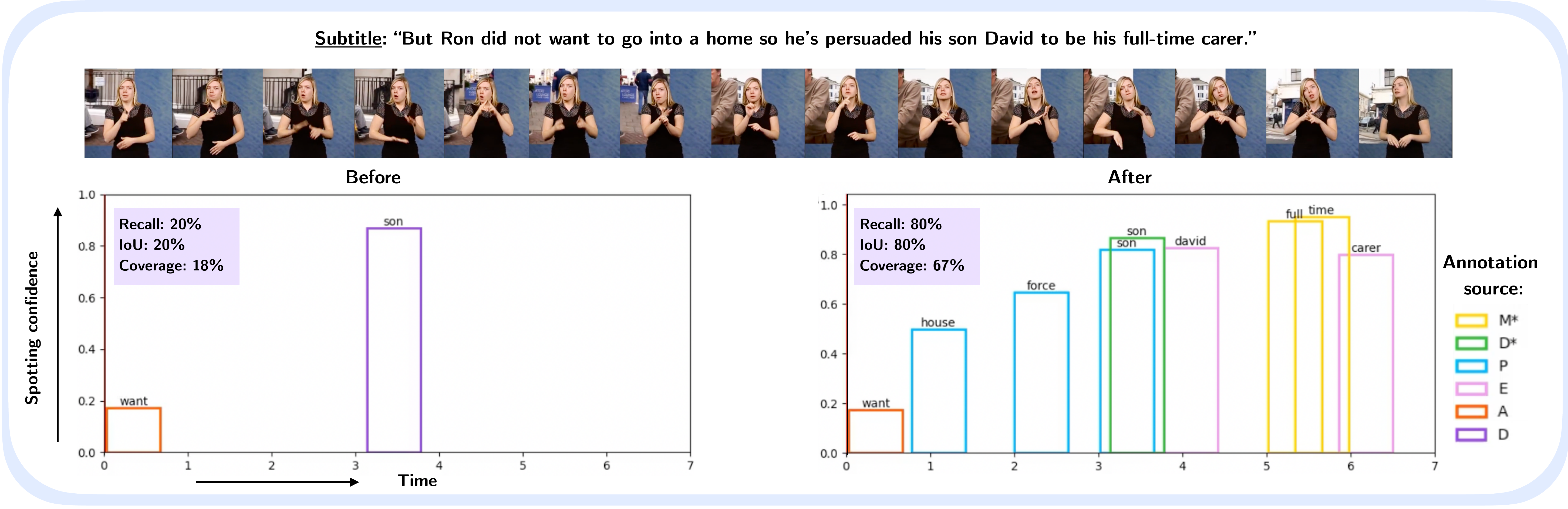}

    \caption{\textbf{Densification}: For continuous sign language, we show automatic sign annotation timelines, along with their confidence and annotation source, \textit{before} and \textit{after} our framework is applied. M, D, A refer to automatic annotations from previous methods from mouthings~\cite{Albanie20}, dictionaries~\cite{Momeni20b} and the Transformer attention~\cite{Varol21}. \spotMnew, \spotDnew, \spotP, \spotExemplarKnown, \spotExemplarNovel{} refer to new and improved automatic annotations collected in this work. Annotation methods are compared in the appendix.
    }
    \label{fig:teaser}

\end{figure}

\section{Introduction}
\label{sec:intro}

Sign languages are visual-spatial languages that have evolved among deaf communities.
They possess rich grammar structures and lexicons that differ considerably from those found among spoken languages~\cite{sutton-spence_woll_1999}.
An important factor impeding progress in automatic sign language recognition -- in contrast to automatic speech recognition -- has been the lack of large-scale training data. To address this issue, researchers have recently made use of sign language interpreted TV broadcasts, comprising (i) a video of continuous signing, and (ii) subtitles corresponding to the audio content, to build datasets such as Content4All~\cite{content4all} (190 hours) and BOBSL~\cite{Albanie2021bobsl} (1460 hours). 

Although such datasets are orders of magnitude larger than the long-standing RWTH-PHOENIX~\cite{camgoz2018neural} benchmark (9 hours) and cover a much wider domain of discourse (not restricted to only weather news), the supervision they provide on the signed content is limited in that it is \textit{weak} and \textit{noisy}. It is weak because the subtitles are temporally aligned with the audio content and not necessarily with the signing. The supervision is also noisy because the presence of a word in the subtitle does not necessarily imply that the word is signed; and subtitles can be signed in different ways. Recent works have shown that training automatic sign language translation models on such \textit{weak} and \textit{noisy} supervision leads to low performance~\cite{content4all,Varol21,Albanie2021bobsl}.

In an attempt to increase the value of such interpreted datasets, multiple works~\cite{Albanie20,Momeni20b,Varol21} have leveraged the subtitles to perform lexical \textit{sign spotting} in an approximately aligned continuous signing segment -- where the aim is to determine \textit{whether} and \textit{when} a subtitle word is signed. Methods include using visual keyword spotting to identify signer mouthings~\cite{Albanie20}, learning a joint embedding with sign language dictionary video clips~\cite{Momeni20b}, and exploiting the attention mechanism of a transformer translation model trained on weak, noisy subtitle-signing pairs~\cite{Varol21}. These works leverage the approximate subtitle timings and subtitle content to significantly reduce the correspondence search space between temporal windows of signs and spoken language words. Although such methods are effective at automatically annotating signs, they only find \textit{sparse} correspondences between keywords in the subtitle and individual signs.

Our goal in this work is to produce \textit{dense} sign annotations, as shown in Fig.~\ref{fig:teaser}. We define densification in two ways: (i) reducing gaps in the timeline so that we have a densely spotted signing sequence; and also (ii) increasing the number of words we recall in the corresponding subtitle. This process can be seen as automatic annotation of lexical signs. Automatic dense annotation of large-vocabulary sign language videos has a large range of applications including: (i)~\textit{substantially} improving recall for retrieval or intelligent fast forwards of online sign language videos; (ii)~enabling \textit{large-scale} linguistic analysis between spoken and signed languages; (iii)~providing \textit{supervision} and \textit{improved alignment} for continuous sign language recognition and translation systems.

In this paper, we ask the following questions: (1) Can we improve current methods to improve the yield of automatic sign annotations whilst maintaining precision? (2) Can we increase the vocabulary of annotated signs over previous methods? (3) Can we `fill in the gaps' that current spotting methods miss? The answer is yes, to all three questions, and we demonstrate this on the recently released BOBSL dataset of British Sign Language (BSL) signer interpreted video. 

\begin{figure}[t]
    \centering
    \includegraphics[width=0.90\textwidth]{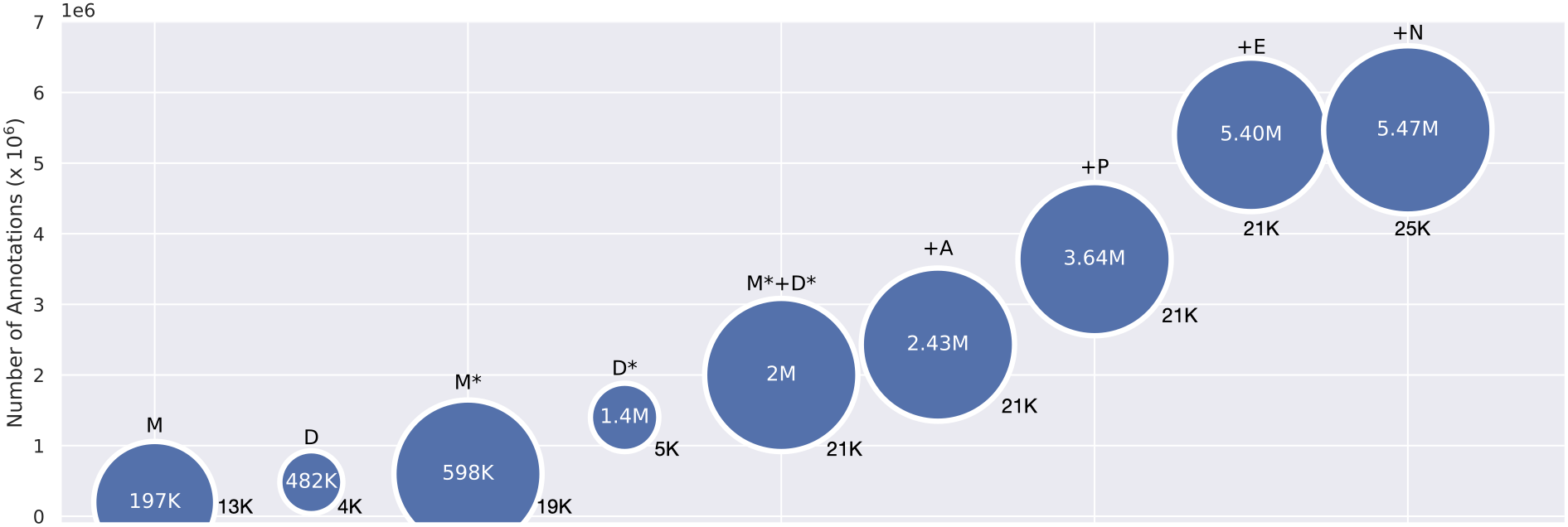}
    \caption{\textbf{Yield of automatic annotations and vocabulary size:} We highlight the increase in the number of automatic annotations and vocabulary size at each stage in our proposed framework. M, D, A refer to annotations from previous methods. \spotMnew, \spotDnew, \spotP, \spotExemplarKnown, \spotExemplarNovel{} refer to new and improved annotations collected in this work. The number of annotations is shown within each circle. The vocabulary size is reported below each circle and also represented by the circle diameter.
    } 
    \label{fig:bar-chart}
\end{figure}

We make the following four contributions: (1)~we significantly improve previous methods by making use of synonyms and subtitle-signing alignment; (2)~we show the value of pseudo-labelling from a sign recognition model as a way of sign spotting; (3)~we propose a novel approach for increasing our annotations of \textit{known} and \textit{unknown} sign classes based on in-domain exemplars; (4)~we will make all 5 million automatic annotations publicly available to support the sign language research community. Our increased yield and vocabulary size is shown in Fig.~\ref{fig:bar-chart}. Our final vocabulary of 24.8K represents the vocabulary of English words (including named entities) from the subtitles which have been automatically associated to a sign instance; different words may have the same sign.

 We note that this work is focused on \textit{interpreted} data, which can differ from \textit{conversational} signing in terms of style, vocabulary and speed~\cite{bragg2019}. Although our long-term aim is to move to conversational signing, learning good representations of signs from interpreted data can be a `stepping stone' in this direction. Moreover, non-lexical signs, such as a pointing sign and spatially located signs, are essential elements of sign language, but our method is limited to the annotation of lexical signs associated to words in the text. 

%% file: 03-related.tex
\section{Related Work}
\label{sec:related}

Our work relates to several themes which we give a brief overview of below. \\

\noindent\textbf{Sign Spotting.} One line of research has focused on the task of \textit{sign spotting},
which seeks to detect signs from a given vocabulary in a target video.
Early efforts for sign spotting employed
lower-level features (colour histograms and geometric cues)
in combination with Conditional Random Fields~\cite{yang2008sign},
Hidden Markov Models (HMMs)~\cite{viitaniemi2014s}
and Sequential Interval Patterns~\cite{ong2014sign}
for temporal modelling. A related body of work has sought to localise signs while leveraging
weak supervision from audio-aligned subtitles. These include the use of external dictionaries~\cite{li2020transferring,Momeni20b,jiang2021looking} and other localisation cues such as
mouthing~\cite{Albanie20} and Transformer attention~\cite{Varol21}.
The performance of these approaches depends on the quality of
the visual features, keywords, and the search window.
In this work, we show improved
yield of existing sign spotting techniques by employing automatic subtitle
alignment techniques to adjust the time window and incorporating synonyms when forming the keywords. Going further beyond the spotting task explored in prior work, we use the automatic spottings to initiate additional algorithms for sign discovery based on \textit{in-domain} exemplar matching (\cref{subsec:exemplars}).
This is similar to dictionary-based sign spotting techniques~\cite{Momeni20b,jiang2021looking}
except we do not source the exemplars from external dictionaries, avoiding the domain gap issue.
Besides in-domain \textit{sign} exemplars as in \cite{jiang2021looking}, we explore the weak
\textit{subtitle} exemplars with unknown sign locations.

A recent progress in mouthing-based keyword spotting was presented
by \textit{Transpotter}~\cite{prajwal2021visual}. This architecture comprises a transformer joint encoder of visual features and phoneme features that is trained to regress both the presence and location of the target keyword in a sequence from mouthing patterns.
Preliminary small-scale experimental results reported by Prajwal et al.~\cite{prajwal2021visual} demonstrated that Transpotter can perform visual keyword spotting in signing footage.
Here, we showcase its suitability for the large-scale annotation regime,
and further train it on sign language data to obtain a greater density
of sign annotations.

In this work, we demonstrate the additional value of \textit{pseudo-labelling}~\cite{yarowsky1995unsupervised,lee2013pseudo} with a sign classifier as an effective mechanism for sign spotting.
While pseudo-labelling has been explored previously for category-agnostic sign segmentation~\cite{renz2021sign} and temporal alignment of glosses~\cite{koller2017re,cheng2020fully} to the best of our knowledge, this is the first use of pseudo-labelling for sign spotting by directly leveraging the predictions of a sign classifier in combination with a pseudo-label filter constructed from the subtitles themselves.

\noindent \textbf{Sign Language Recognition.}
Efforts to develop visual systems for sign recognition stretch
back to work in 1988 from Tamaura and Kawasaki~\cite{tamura1988recognition},
who sought to classify signs from hand location and motion features.
There were later efforts to design hand-crafted features
for sign recognition~\cite{charayaphan1992image,starner1995visual,vogler1997adapting,vogler1998asl,ong2012sign}. Deep convolutional neural networks then came to dominate sign representation~\cite{koller2016deep},
particularly via 3D convolutional architectures~\cite{joze2018ms,li2020word,Albanie20,li2020transferring} with extensions to focus model capacity around human skeletons~\cite{huang2018attention} and non-manual features~\cite{hu2021global}.

In the domain of continuous sign language recognition,
in which the objective is to infer a sequence of sign glosses,
prior work has explored HMMs~\cite{bauer2000relevant,koller2015continuous} in combination with Dynamic Time Warping (DTW)~\cite{zhang2014threshold},
RNNs~\cite{cui2017recurrent}
and architectures capable of learning effectively from CTC losses~\cite{zhou2020spatial,cheng2020fully}. 
Recently, sign representation learning methods inspired by BERT~\cite{devlin2018bert} have shown the potential to learn effective representations for both isolated~\cite{hu2021signbert} and continuous~\cite{zhou2021signbert} recognition.
Koller~\cite{Koller2020QuantitativeSO} provides an extensive survey of the sign recognition literature, highlighting the extremely limited supply of datasets with large-scale vocabularies suitable for continuous sign language recognition.
In our work, we aim to take a step towards addressing this gap by developing ``densification'' techniques for constructing such datasets automatically.

\noindent \textbf{Sign Language Translation.}
The task of translating sign language video to spoken language sentences was first tackled with neural machine translation by Camg\"oz et al.~\cite{camgoz2018neural}, who also introduced the PHOENIX-Weather-2014T dataset to facilitate research on this topic.
Several frameworks have been proposed to employ transformers for this task~\cite{camgoz2020sign,Yin2020BetterSL}, with extensions to improve temporal modelling~\cite{li2020tspnet}, multi-channel cues~\cite{Camgoz2020MultichannelTF} and signer independence~\cite{Jin2021ContrastiveDM}.
Related work has also sought to contribute to progress on this task by exploiting monolingual data~\cite{zhou2021improving} and gloss sequence synthesis~\cite{Moryossef2021DataAF,li2021transcribing}.
To date, various works have shown promise on the PHOENIX-Weather 2014T~\cite{camgoz2018neural} and CSL Daily~\cite{zhou2021improving} benchmarks.
However, sign language translation has not yet been demonstrated for a large vocabulary across multiple domains of discourse.
Differently from the works above, this paper focuses on developing methods that are applicable to large/open vocabulary regimes.

\noindent\textbf{Weakly-supervised Object Discovery and Localisation.}
Our approach is also related to the rich body of literature on object cosegmentation~\cite{rother2006cosegmentation,joulin2010discriminative,kim2011distributed,rubinstein2013unsupervised}, weakly supervised object localisation~\cite{nguyen2009weakly,deselaers2010localizing,shi2013bayesian,wang2014weakly,gokberk2014multi}, object colocalisation~\cite{tang2014co,joulin2014efficient} and unsupervised object discovery and localisation~\cite{cho2015unsupervised,vo2021large}.
Here, we propose an algorithm for discovering and localising novel signs (i.e.\ for which we have no labelled examples), but instead have weak supervision in the form of subtitles containing keywords of interest.
Moving beyond initial work that sought to learn from subtitles in an aligned setting~\cite{farhadi2006aligning}, classical approaches for sign discovery using subtitles have included Multiple Instance Learning where the subtitles are considered as positive and negative bags for a particular keyword~\cite{buehler2009learning,kelly2010weakly,pfister2013large} and a priori mining~\cite{cooper2009learning}.
Differently from these works, we first bootstrap our sign discovery process with sign spotting to both obtain initial candidates and learn robust sign representations, then propagate these examples across video data by leveraging the similarities between the resulting representations together with noisy constraints imposed by the subtitle content.

%% file: 04-method.tex
\section{Densification}
\label{sec:method1}
Our goal is to leverage several ways of sign spotting to achieve dense
annotation on continuous signing data. To this end, we introduce
both new sources of automatic annotations, and also improve the existing sign spotting
techniques. We start by presenting two new spotting methods using in-domain exemplars: to mine more sign instances with individual \textit{exemplar signs} (Sec.~\ref{subsec:exemplars})
and to discover novel signs with weak \textit{exemplar subtitles} (Sec.~\ref{subsec:novel}). We also show the value of pseudo-labelling from a sign recognition model for sign spotting (Sec.~\ref{subsec:pl}). We then describe key improvements to previous work which substantially increase the yield of automatic annotations (Sec.~\ref{subsec:improvements}). Finally, we present our evaluation framework to measure the quality of our sign spottings in a large-vocabulary setting (Sec.~\ref{subsec:framework}). The contributions of each source of annotation are assessed in the experimental results. 

\subsection{Mining more Spottings through In-domain Exemplars (\spotExemplarKnown)}
\label{subsec:exemplars}

The key idea is: given a continuous signing video clip and a set of exemplar clips of a particular sign, we can use the exemplars to search for that sign within the video clip. In our case, the exemplars are obtained from other \textit{automatic} spotting methods (\spotMnew, \spotDnew, A, \spotP), described in Sec.~\ref{subsec:pl} and Sec.~\ref{subsec:improvements}, and come from the same \textit{domain} of sign language interpreted data, i.e. the same training set. We hypothesise that signs from the same domain are more likely to be signed in a similar way and in turn help recognition; in contrast, for example, to signs from a different domain such as dictionaries.

Formally, suppose we have a reference video $V_{0}$ in which we wish to localise a particular sign $w$, whose corresponding word occurs in the subtitle. We also have $N$ video exemplars $V_1, \dots, V_N$ of the sign $w$. For each video, $V_i$, let $\mathcal{C}_i$ denote the set of possible temporal locations of the sign $w$ and let $c = (f, p) \in \mathcal{C}_i$ denote a candidate with features $f$ at temporal location $p$. We compute a score map between our reference video $V_{0}$ and each exemplar $V_1, \dots, V_N$ by computing the cosine similarity between each feature at each position in $c_0 \in \mathcal{C}_0$ and $(c_1, c_2, \dots c_n) \in \mathcal{C}_1 \times \dots \times \mathcal{C}_N$. This results in $N$ score maps of dimension $|\mathcal{C}_0| \times |\mathcal{C}_i|$ for $i=1\dots N$. We then apply a max operation over the temporal dimension of the exemplars, giving us $N$ vectors of length $|\mathcal{C}_0|$, which we call ${M}_1, \dots, {M}_N$. 

We subsequently apply a voting scheme to find the location of the common sign $w$ in $V_{0}$. Specifically, we let $L=\frac{1}{N}\sum_{i=1}^{N} \mathbbm{1}_{({M}_i>h)}$ for a threshold $h$, where the vector $\mathbbm{1}_{(M_i>h)}$ takes the value 1 for entries of $M_i$ which are greater than $h$ and 0 otherwise. 
The candidate location of $w$ in $V_{0}$ is then $c = (f, p) \in \mathcal{C}_0$ where $p$ corresponds to the position of the maximum non-zero entry in the vector $L$
(see Fig.~\ref{fig:exemplars} for a visual illustration).
If there are multiple maxima, we assign $p$ to be the midpoint of the largest connected component. If all entries in $L$ are zero, we conclude $w$ is not present. We perform two variants of this approach using mean and max pooling of the score maps (instead of voting); these are described in the appendix. We note that for a given signing sequence, we only focus on finding signs for words in the subtitle that have \textit{not} been annotated by other methods.

\subsection{Discovering Novel Sign Classes (\spotExemplarNovel)}
\label{subsec:novel}

\begin{figure}[t]
    \centering
    \includegraphics[width=0.9\textwidth]{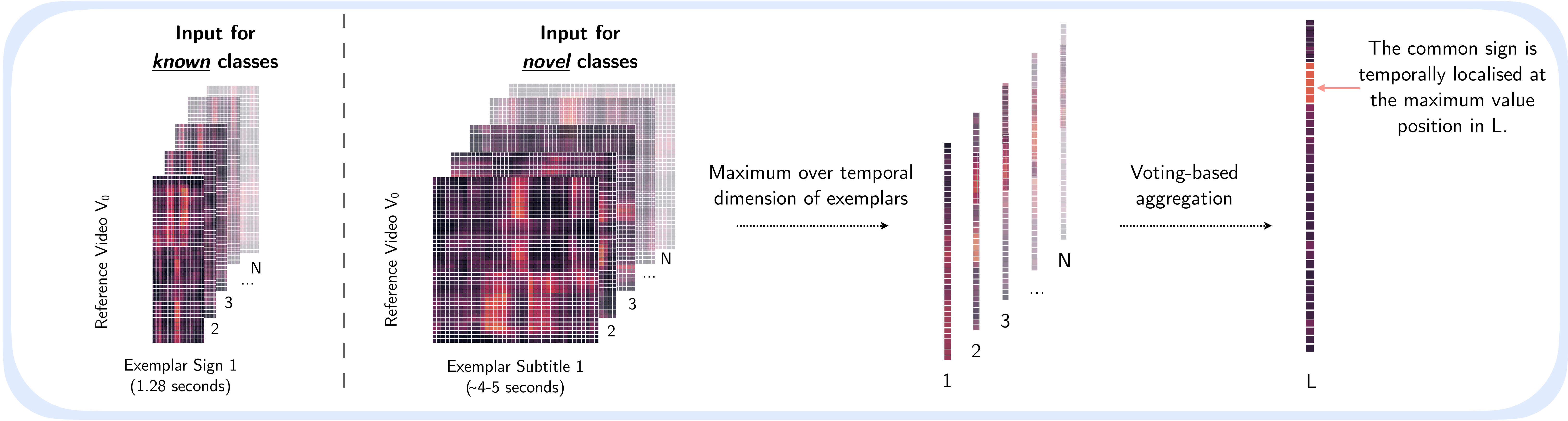}
    \caption{\textbf{Sign spotting through exemplars to find instances of known classes (E) and novel classes (N)}: By comparing a reference video $V_0$ to a set of exemplars (either sign exemplars for known sign class instances or weak subtitle exemplars for novel sign class instances), we can find the common lexical sign in the collection. We (1) form a set of score maps by calculating the cosine similarities between reference and exemplar representations; (2) we perform a maximum operation over the temporal dimension of exemplars; (3) we apply a voting-based aggregation to find the temporal location of the commmon sign in $V_0$. The duration of exemplar signs is fixed. 
    }
    \label{fig:exemplars}
\end{figure}

One limitation of our proposed method in Sec.~\ref{subsec:exemplars} is that we are only able to collect more sign instances from a \textit{closed} vocabulary, determined by sign exemplars obtained from other methods (described in Sec.~\ref{subsec:pl} and Sec.~\ref{subsec:improvements}). Here, we extend our approach to localise \textit{novel} signs, for which we have no exemplar signs but whose corresponding word appears in the subtitle text. We follow our approach described in Sec.~\ref{subsec:exemplars}, computing score maps between our reference video and exemplar subtitles (instead of exemplar signs,
see Fig.~\ref{fig:exemplars}). We note that by `exemplar subtitle', we are referring to the video frames corresponding to the subtitle timestamps. Non-lexical signs, such as pointing signs or pause gestures, are very common in sign language. To avoid annotating such non-lexical signs as the common sign across $V_0$ and  $V_1, \dots, V_N$, we also choose $N^-$ negative subtitle exemplars ${U}_1 \dots {U}_{N^-}$ presumed to not contain $w$ (due to the absence of $w$ in the subtitle). We compute $L^+$ and $L^-$ using the score maps from positive exemplars $V_1, \dots, V_N$ and negative exemplars ${U}_1, \dots, {U}_{N^-}$ respectively. We then let $L=L^+ - L^-$. Implementation details on the number of positive and negative exemplars used can be found in the appendix.

\subsection{Pseudo-labelling as a Form of Sign Spotting (\spotP)}
\label{subsec:pl}

We propose to re-purpose a pretrained large-vocabulary sign classification model
(see vocabulary expansion in Sec.~\ref{subsec:framework})
for the task of sign spotting.
Specifically, we predict a sign class from a fixed vocabulary for each time step in a continuous signing video clip. We subsequently filter the predicted signs to words which occur in the corresponding English subtitle. Similarly to~\cite{Varol21}, here the task is to recognise the sign from scratch, without a query keyword. The subtitle is only used as a post-processing step to filter out signs which are less likely performed (due
to absence in the subtitle).
 
\subsection{Improving the Old (\spotMnew, \spotDnew)}\label{subsec:improvements}

Here, we briefly describe our improvements over the existing sign spotting techniques, additional details are provided in the appendix.

\noindent\textbf{Better Mouthings with an Upgraded KWS from Transpotter~\cite{prajwal2021visual}.} In previous work~\cite{Albanie2021bobsl}, an improved BiLSTM-based visual-only keyword spotting model of Stafylakis et al.~\cite{stafylakis2018zero} from~\cite{momeni2020seeing} (named ``P2G~\cite{stafylakis2018zero} baseline") is used to automatically annotate signs via mouthings. In this work, we make use of the recently proposed transformer-based \textit{Transpotter} architecture~\cite{prajwal2021visual}, provided by the authors, that achieves state-of-the-art results in visual keyword spotting on lipreading datasets. We follow the procedure described in~\cite{Albanie20,Albanie2021bobsl} to query words in the subtitle in continuous signing video clips.

\noindent\textbf{Finetuning KWS on Sign Language Data through Bootstrapping.} The visual keyword spotting Transpotter architecture in~\cite{prajwal2021visual} is trained on silent speech segments, which differ considerably from signer mouthings. In fact, signers do not mouth continuously and sometimes only partly mouth words~\cite{Boyes2001mouthing}. In order to reduce this severe domain gap, we propose a dual-stage finetuning strategy. First, we extract high-confidence mouthing annotations using the pre-trained Transpotter from~\cite{prajwal2021visual} on the BOBSL training data. We query for the words in the subtitle and obtain the temporal localization of the word in the video. We finetune on this pseudo-labeled data using the same training pipeline of~\cite{prajwal2021visual}, where the spotted mouthings (word-video pairs) act as positive samples. For the negative samples, we pair a given word with a randomly sampled video segment from the dataset. As we observe the Transpotter to predict a large number of false positives, we remedy this by sampling a larger number of negative pairs in each batch. We also do a second round of fine-tuning by training on the pseudo-labels from the finetuned model of the first stage. We did not achieve significant improvements with further iterations. 

\noindent\textbf{Better Search Window with Subtitle Alignment with SAT~\cite{bull2021}.}
One challenge in using sign language interpreted TV broadcasts is that the original subtitles are not aligned to the signing, but to the audio track. In \cite{Albanie2021bobsl}, a signing query window is defined
as the audio-aligned subtitle timings together with padding on both sides to account
for the misalignment. We automatically align spoken language text subtitles to the signing video by using the SAT model introduced in~\cite{bull2021}, trained on manually aligned and
pseudo-labelled subtitles as described in~\cite{Albanie2021bobsl}.
By using subtitles which are better aligned to the signing,
we reduce the probability of missing spottings.

\noindent\textbf{Better Keywords with Synonyms and Similar Words.}
To determine whether a keyword belongs to a subtitle, previous works
\cite{Albanie2021bobsl}
check whether the raw form, the lemmatised form,
or the text normalised form (e.g.\ \textit{two} instead of \textit{2})
appears in the subtitle text. We notice that this is sub-optimal
as multiple words may correspond to the same sign, often due to
(i)~English synonyms, (ii)~identical signs for similar words, or (iii)~ambiguities in spoken language. For example, \emph{dad} and \emph{father} or \emph{today} and \emph{now} can be the same signs in BSL. 
In this work, we investigate whether the automatic annotation
yield could be improved by querying words beyond the subtitle, by querying synonyms and similar words to the words in the subtitle. We collect the additional words to query through (i)~English synonyms from WordNet~\cite{wordnet}, (ii)~the metadata
present in online sign language dictionaries such as SignBSL\footnote{\url{www.signbsl.com}}~\cite{Momeni20b} and BSL Sign-Bank\footnote{\url{bslsignbank.ucl.ac.uk}}  
which provide a set of `related words' for each sign video entry; (iii)~words with GloVe~\cite{pennington-etal-2014-glove} cosine similarity above 0.9 to account for ambiguities in spoken language.

\subsection{Evaluation Framework}
\label{subsec:framework}

\begin{figure}[t]
    \centering
    \includegraphics[width=0.9\textwidth]{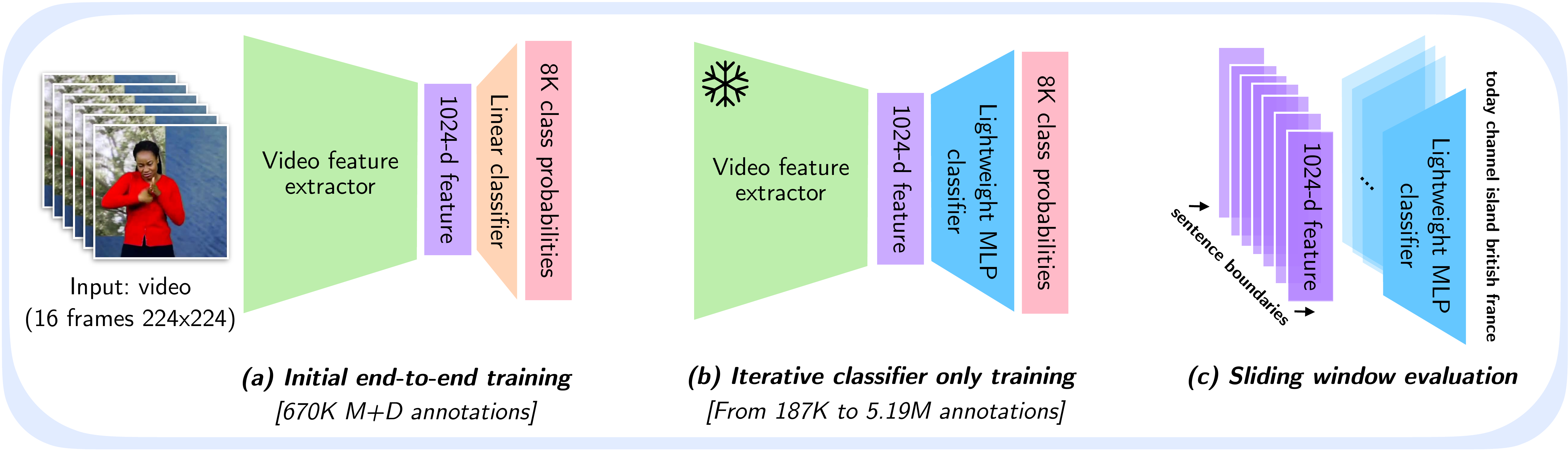}
    \caption{\textbf{Evaluation framework}:
    (a) Video features are obtained by training an I3D architecture end-to-end given M + D annotations from~\cite{Albanie2021bobsl}. The I3D ingests  16-frames of video and has a linear classifier for 8K sign categories.
    The end-to-end training is a costly procedure which is not  affordable to repeat
    for each set of our new sign spottings that are on the order of several million training samples.
    (b) As new sets of spottings are generated, a light weight MLP classifier is trained on the pre-extracted I3D features. This relatively inexpensive training procedure means that we benefit from new annotations without the expense of end-to-end training. (c) The MLP is applied in a sliding window fashion to the signing sequence to generate sign predictions.
    }
    \label{fig:framework}
\end{figure}

Our framework consists of three stages: (a)~a costly end-to-end classification training
to learn sign category aware video features given an initial set of sign-clip annotation pairs;
(b)~a lightweight classification training given pre-extracted video features
for a large number of annotations; (c)~a sliding window evaluation of the trained lightweight model by comparing dense sign predictions against the subtitles
(see Sec.~\ref{subsec:evaluation}). These stages are illustrated in Fig.~\ref{fig:framework}.
Note that the \textit{annotations} we refer to are
always \textit{automatically} localised sign spottings from continuous videos using subtitle information.
The motivation for the video backbone and lightweight classifier is purely related to computational costs. Unlike traditional
video recognition datasets, we work with untrimmed video data of 1400 hours, where the
set of sign-clip pairs is not fixed. Instead, our goal is to increase the number of sign-clip
pairs within the continuous stream, and assess the quality of the expanded annotation yield
on the proxy task of continuous sign language recognition.
Next, we describe the training stages for the video backbone and the lightweight classifier.

\noindent\textbf{Improving the I3D Feature Extractor through Vocabulary Expansion.}
Following previous works~\cite{joze2018ms,li2020word,Albanie20,Albanie2021bobsl},
we use the I3D spatio-temporal convolutional architecture
to train an end-to-end sign recognition model.
We input 16 consecutive RGB frames and output class probabilities.
The details about optimisation are provided in the appendix.
As explained above, this model
forms the basis of sign video representation which corresponds to the
spatio-temporally pooled latent embedding before the classification layer.
The prior work of \cite{Albanie2021bobsl} trains this classifier on the BOBSL dataset (see Sec.~\ref{subsec:evaluation}) with 2K categories 
obtained through the vocabulary of mouthing spottings.
As a first step, we perform a vocabulary expansion and construct a significantly
increased vocabulary of 8K categories. This is achieved by including each sign
that has at least 5 training spottings above 0.7 confidence from both mouthing (M) and dictionary (D) 
annotations. The confidence for the mouthing annotation corresponds to the probability that a text keyword (corresponding to the sign) is mouthed at a certain time frame, as computed in~\cite{Albanie20}. The confidence for the dictionary annotation corresponds to the cosine similarity (normalised between 0-1) between the representations of a dictionary clip of the sign and the continuous signing at each time frame, as in~\cite{Momeni20b}. The resulting M+D training
set comprises 670K annotations, with a long-tailed distribution. Furthermore,
we note that the categories are noisy where multiple categories may correspond
to the same sign, and vice versa. Despite this noise, we empirically show
that this model provides better performance than its 2K-vocabulary counterpart.
We use our improved I3D model for two purposes: as the frozen feature extractor
and as the source of pseudo-labelling for sign spotting (see Sec.~\ref{subsec:pl}).

\noindent\textbf{Lightweight Sign Recognition Model.}
Following \cite{Varol21}, we opt for a 4-layer MLP module (with one residual connection) to assess the quality of
different sets of annotations.
Given pre-extracted features, this model is trained
for sign recognition into 8K categories. We note that we do not train on a larger vocabulary to avoid the presence of many singletons in the training set. The efficiency of the MLP allows faster experimentation
to analyse the value of each of our sign spotting sets.
The input is one randomly sampled feature around the sign spotting location (the receptive
field of one feature 16 frames). The MLP weights are randomly initialised.
Additional training and implementation details are given in the appendix
\if\sepappendix1{Sec.~B.5}
\else{Sec.~\ref{app:subsec:i3d}}
\fi
and
\if\sepappendix1{B.6.}
\else{\ref{app:subsec:mlp}.}
\fi

%% file: 05-experiments.tex
\section{Experiments}
\label{sec:experiments}

We start by describing our dataset and evaluation metrics (Sec.~\ref{subsec:evaluation}). We then present experimental results on the contribution of each source of annotation and show qualitative examples (Sec.~\ref{subsec:results}).

\subsection{Data and Evaluation Protocol}
\label{subsec:evaluation}

\noindent \textbf{BOBSL}~\cite{Albanie2021bobsl} is a public dataset consisting of British Sign Language interpreted BBC broadcast footage, along with English subtitles corresponding to the audio content. The data contains 1,962 episodes, which have a total duration of 1,467 hours spanning 426 different TV shows. BOBSL has a total 1,193K subtitles covering a total vocabulary of 78K words. We note that in this work we use the word \textit{subtitle} to refer to the processed BOBSL sentences from~\cite{Albanie2021bobsl} as opposed to the raw subtitles. There are a total of 39 signers in the dataset. Further dataset statistics can be found in~\cite{Albanie2021bobsl}. For a subset of 36 episodes in BOBSL, referred to as SENT-TEST in~\cite{Albanie2021bobsl}, the English subtitles have been manually aligned \textit{temporally} to the continuous signing video. We make use of this test set to evaluate the quality of our predicted automatic annotations. SENT-TEST covers a total duration of 31 hours and contains 20,870 English subtitles. The total vocabulary of English words is 13,641, of which 5,604 are singletons. The 3 signers in SENT-TEST are different to the signers in the training set, this enables signer-independent BSL recognition to be evaluated.\\

\noindent \textbf{Evaluation protocol.} Given an English subtitle and the \textit{temporally} aligned continuous signing video clip, we evaluate our  predicted signs for the clip using (i) \textit{intersection over union} (IoU); (ii) \textit{recall} between signs and the English word sequence; and (iii) \textit{temporal coverage}: this is defined as the proportion of frames in the clip assigned to signs that occur in the word sequence, where a sign is given a fixed duration of 16 frames (for 25Hz video). Note that none of these metrics depend on the word order of the English subtitle, only the words it contains. All metrics are rescaled from the range 0-1 to 0-100 percentage for readability.

\begin{figure}[t]
    \centering
    \includegraphics[width=0.9\textwidth]{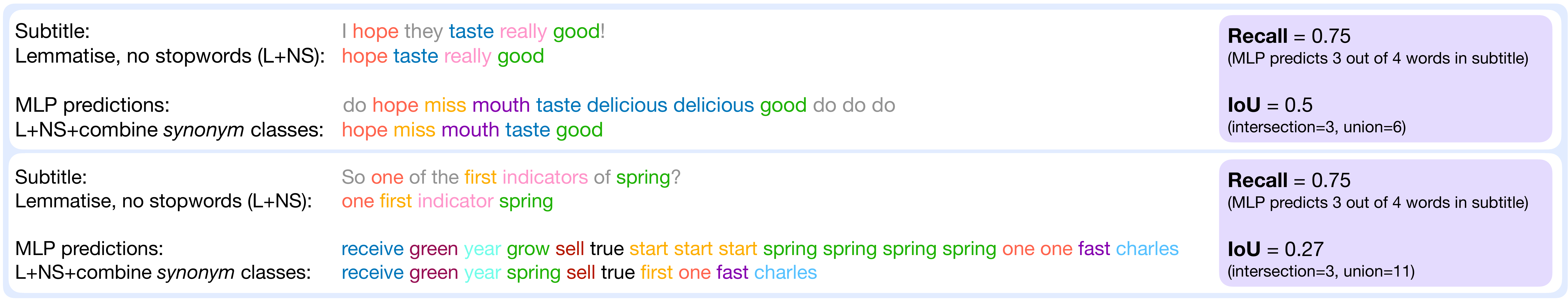}
   \caption{
    \textbf{Evaluation illustration on sample prediction: }We illustrate the processing applied to the predicted sign sequence from the MLP predictions and corresponding English subtitle for calculating our metrics. As the MLP model predicts one sign per time-step, some predictions are repeated and irrelevant words appear at transition periods between signs, decreasing the IoU. Some signs are not predicted as they are not signed, showing the limitations of using the subtitle to measure performance.
    \label{fig:eval-protocol}
    }
\end{figure}

For this evaluation, stop words are filtered out since often they are not signed. This reduces the number of test subtitles from 20,870 to 20,547: subtitles such as ``is it?'', ``Oh!'', ``but no'' are removed. The sign and word sequences are also lemmatised. We also remove repetitions from the predicted sign sequence and allow the prediction of synonyms of words in the English subtitle. This processing is highlighted in Fig.~\ref{fig:eval-protocol}, where the IoU and recall are computed for a pair of predicted signs and English text. 
While this evaluation is suboptimal due to the simplified
word-sign correspondence assumption, it tests the capacity of the sign recognition
model in a large-vocabulary scenario, necessary for open-vocabulary sign language technologies.

Note, the predicted signs for a clip can be produced 
in two ways. In the first way, the signs are obtained from the automatic annotations using knowledge of the content of the English subtitle -- we refer to these as {\em Spottings}.
In the second, signs are predicted directly from the clip using the MLP sign predictions, without access to the corresponding English subtitle. These are referred to as {\em MLP predictions}.
Spottings are evaluated using all the words; this metric is important to monitor how dense we can
automatically annotate the data. The MLP evaluation is limited to the fixed classification vocabulary (of size 8K in our experiments). We note that when different annotations are combined, the sign spotting methods are applied independently.

\subsection{Results}\label{subsec:results}

\noindent\textbf{Comparison of Video Features.}
By finetuning our Kinetics pretrained I3D model on BOBSL M+D annotations from~\cite{Albanie2021bobsl} using an 8K vocabulary instead of a 2K vocabulary, we improve predictions on the test set, as shown in~\cref{tab:i3d-features}. We increase the recall from 25.5 to 26.3 and the coverage from 15.5 to 16.3. We therefore use the 8K M+D model for the rest our experiments as the frozen feature extractor. We note that we restrict the M+D annotations to the high-confidence ones (over 0.8 threshold) used for the I3D baseline in~\cite{Albanie2021bobsl}, as these present an appropriate signal-to-noise ratio. We use the same threshold for subsequent automatic annotations unless stated otherwise. 

\begin{table}[t]
    \centering
     \caption{\textbf{Comparison of I3D video features}: We highlight the improved performance of I3D on the test set (SENT-TEST) when trained on a larger vocabulary (8K instead of 2K) with more samples (670K instead of 426K).}
    \setlength{\tabcolsep}{25pt}
    \resizebox{\linewidth}{!}{
    \begin{tabular}{l|c|c|c|c|c}
        \toprule
         &  &  &\multicolumn{3}{|c}{\textbf{I3D predictions}} \\
          &  &  &\multicolumn{3}{|c}{\textbf{(subtitle independent)}} \\
        Annot. source & Num. I3D train annot. & Vocab. size & Recall & IoU & Coverage \\
        \midrule

       M~\cite{Albanie20}+D~\cite{Momeni20b}  & 426K & 2K & 25.5 & 6.4 & 15.5\\
         M~\cite{Albanie20}+D~\cite{Momeni20b}  & 670K & 8K&\textbf{26.3} & \textbf{7.9 }&\textbf{16.3} \\
        \bottomrule
    \end{tabular}
    }
    \label{tab:i3d-features}
\end{table}

\begin{table}[t]
    \centering
    \caption{\textbf{Improved mouthing and dictionary spottings:}
    We evaluate different sets of spottings and their respective MLP predictions.
    M~\cite{prajwal2021visual} shows our finetuned version for all the rows in the last block.
    We quantify the effects of subtitle alignment and querying synonyms.
    We also show the oracle performance and a translation baseline.} 
    \setlength{\tabcolsep}{4pt}
    \resizebox{\linewidth}{!}{
    \begin{tabular}{l|c|c|rrr|ccc|ccc}
        \toprule
         & &  & \multicolumn{3}{|c|}{\textbf{Training set}} &\multicolumn{3}{|c|}{\textbf{Spottings [full]}}& \multicolumn{3}{|c}{\textbf{MLAP predictions [8K]}} \\
         & Subtitle &  & \multicolumn{1}{c}{full} & \#ann. & \#ann.&\multicolumn{3}{|c|}{\textbf{(subtitle dependent)}} & \multicolumn{3}{|c}{\textbf{(subtitle independent)}} \\
        Annotation source  & alignment & Synonyms & vocab & [full] & [8K] & Recall & IoU & Coverage & Recall & IoU & Coverage\\
        \midrule  

        Oracle                                   &            &                    &   -    &  -     &  -    &     - &   -   &  -     & 86.7 & 86.3 & 55.2 \\
        \midrule  
        Translation baseline~\cite{Albanie2021bobsl}        &            &                    &   -    &      - &    -  &    -  &   -   &    -   & 11.7 & 8.3 &  7.6 \\
        \midrule
        M~\cite{Albanie20}                                  &            &                    & 13.6K &  197K &  187K &  2.5 &  2.2 &  1.3 & 15.1 & 3.2 &  8.7 \\
        M~\cite{prajwal2021visual} (no finetuning)          &            &                    &  21.5K     &    725K   &    661K  &  9.4    &   8.3   & 4.9      & 20.4 & 4.8 & 11.9 \\
        \midrule
        M~\cite{prajwal2021visual}                          &            &                    & 18.6K &  445K &  412K &  7.1 &  6.5 &  3.9 & 23.6 & 4.8 & 13.8 \\
        M~\cite{prajwal2021visual}  (\spotMnew)             & \checkmark &                    & 19.6K &  598K &  552K &  8.9 &  8.2 &  4.9 & 27.4 & 6.3 & 16.7 \\
        M~\cite{prajwal2021visual}                          & \checkmark & \checkmark         & 19.6K & 1.38M & 1.25M & 11.8 & 10.4 &  6.1 & 25.3 & 6.2 & 16.3 \\
        D~\cite{Momeni20b}                                  &            &                    &  4.4K &  482K &  482K &  6.5 &  6.3 &  3.7 & 24.0 & 7.2 & 15.1 \\
        D~\cite{Momeni20b}                                  & \checkmark &                    &  4.5K &  535K &  535K &  7.0 &  6.9 &  4.0 & 24.2 & 7.3 & 15.3 \\
        D~\cite{Momeni20b} (\spotDnew)                      & \checkmark & \checkmark         &  5.0K & 1.40M & 1.39M & 12.5 & 11.6 &  7.0 & 26.0 & 7.3 & 16.9 \\
        \spotMnew + \spotDnew                               & \checkmark & \checkmark(D-only) & 20.9K & 2.00M & 1.94M & 19.0 & 17.6 & 10.5 & 29.0 & 7.9 & 18.4 \\
        \spotMnew + \spotDnew + A~\cite{Varol21}            & \checkmark & \checkmark(D-only) & 20.9K & 2.43M & 2.37M & \textbf{21.9} &\textbf{20.1} &\textbf{11.8} & \textbf{29.6}& \textbf{9.1} & \textbf{19.0} \\

        \bottomrule
    \end{tabular}
    }
    
    \label{tab:improved-mouthing-dict-pl}
\end{table}

\noindent\textbf{Oracle.}
As the MLPs are trained on a restricted 8K vocabulary, it is not possible to predict the full vocabulary of 13,641 words present in the test set subtitles. Furthermore, not all words in the subtitle are signed and vice versa. This means a recall, IoU and coverage of 100\% is not achievable between predicted signs and English subtitle words. However, we propose an oracle in Tab.~\ref{tab:improved-mouthing-dict-pl} whereby we measure the recall and IoU assuming each word in the subtitle, which either falls within the 8K vocabulary or corresponds to a synonym of a word in the 8K vocabulary, is signed and correctly predicted. The oracle achieves a recall of 86.7 and IoU of 86.3. For the coverage metric, we assume each correctly predicted sign has a duration of 16 frames and no signs overlap. The resulting oracle coverage is 55.2. This low coverage is partly due to the signer pausing within subtitles and also due to the presence of non-lexical signs. In fact, the percentage of fully lexical signs in three other sign language corpora (Auslan~\cite{johnston2012lexical}, ASL~\cite{johnston2012lexical} and LSF~\cite{belissen2020experimenting}) is estimated to be only 70-85\% of total signing.

\noindent\textbf{Translation Baseline.} Although the goal in this work is not translation, but achieving dense annotations, we can nevertheless compare our MLP predictions to the translation baseline in~\cite{Albanie2021bobsl}. Using the test set translation predictions from this model, we perform the same processing as highlighted in Fig.~\ref{fig:eval-protocol} to calculate our metrics. As shown in Tab.~\ref{tab:improved-mouthing-dict-pl}, all our simple MLP models clearly outperform the transformer-based translation model used in~\cite{Albanie2021bobsl}, demonstrating that we are able to recognise more signs in the English subtitle.

\noindent\textbf{Improving Mouthing and Dictionary Spottings.} As shown in Tab.~\ref{tab:improved-mouthing-dict-pl}, by using the Transpotter~\cite{prajwal2021visual} for spotting mouthings M, our yield of total annotations triples from 197K to 725K. The quality of these new annotations is reflected in the increased performance of the MLP: the recall increases from 15.1 to 20.4 and the coverage from 8.7 to 11.9. Finetuning the keyword spotter on sign language data through pseudo-labelling also helps considerably despite the drop in the number of training annotations since there are less false positives; recall increases from 20.4 to 23.6 and coverage from 11.9 to 13.8. Subtitle alignment improves the yield of both mouthing and dictionary annotations, as shown in Tab.~\ref{tab:improved-mouthing-dict-pl}. This translates to a significant boost for mouthings on the MLP performance; the recall increases from 23.6 to 27.4 and the coverage from 13.8 to 16.7. For dictionary annotations, the improvement by using aligned subtitles is less striking. By querying synonyms when searching for mouthings, the yield more than doubles. However, these additional annotations seem to be quite noisy as they decrease the performance of our MLP. Due to the nature of sign language interpretation, it is possible that signers are far more likely to mouth a word which is actually in the written subtitle than a synonym of that word. We therefore do not query synonyms for mouthing spottings. For dictionary spottings, we observe the opposite effect.  By incorporating synonyms, the yield of dictionary spottings more than doubles and the recall of the MLP predictions also increases from 24.2 to 26.0. We denote our best performing mouthing and dictionary spottings with \spotMnew{} and \spotDnew{}, respectively.
Adding attention spottings from~\cite{Albanie2021bobsl} (with a threshold of 0) adds around 400K additional annotations and boosts the MLP performance; increasing recall from 29.0 to 29.6 and coverage from 18.4 to 19.0, compared to the oracle recall of 86.7 and coverage of 55.2.

\noindent{\textbf{Sign Recognition as a Form of Pseudo-labelling.}} 
Pseudo-labels \spotP{} are a source of over 1M new annotations (when using a threshold of 0.5) on top of our best \spotMnew, \spotDnew, A spottings. As shown in Tab.~\ref{tab:exemplars_with_novel}, they greatly increase the spottings recall from 21.9 to 25.4 and coverage from 11.8 to 13.9, while only marginally increasing the recall and coverage for MLP predictions. As the pseudo-labels come from our 8K I3D model in Tab.~\ref{tab:i3d-features} whose frozen features are also used for training the MLP, \spotP{} may not be providing additional information for our downstream evaluation. Nevertheless, they provide a great source of additional spottings (not found by previous methods) for our goal of dense annotation.

\begin{figure}[t]
    \centering
    \includegraphics[width=.90\textwidth]{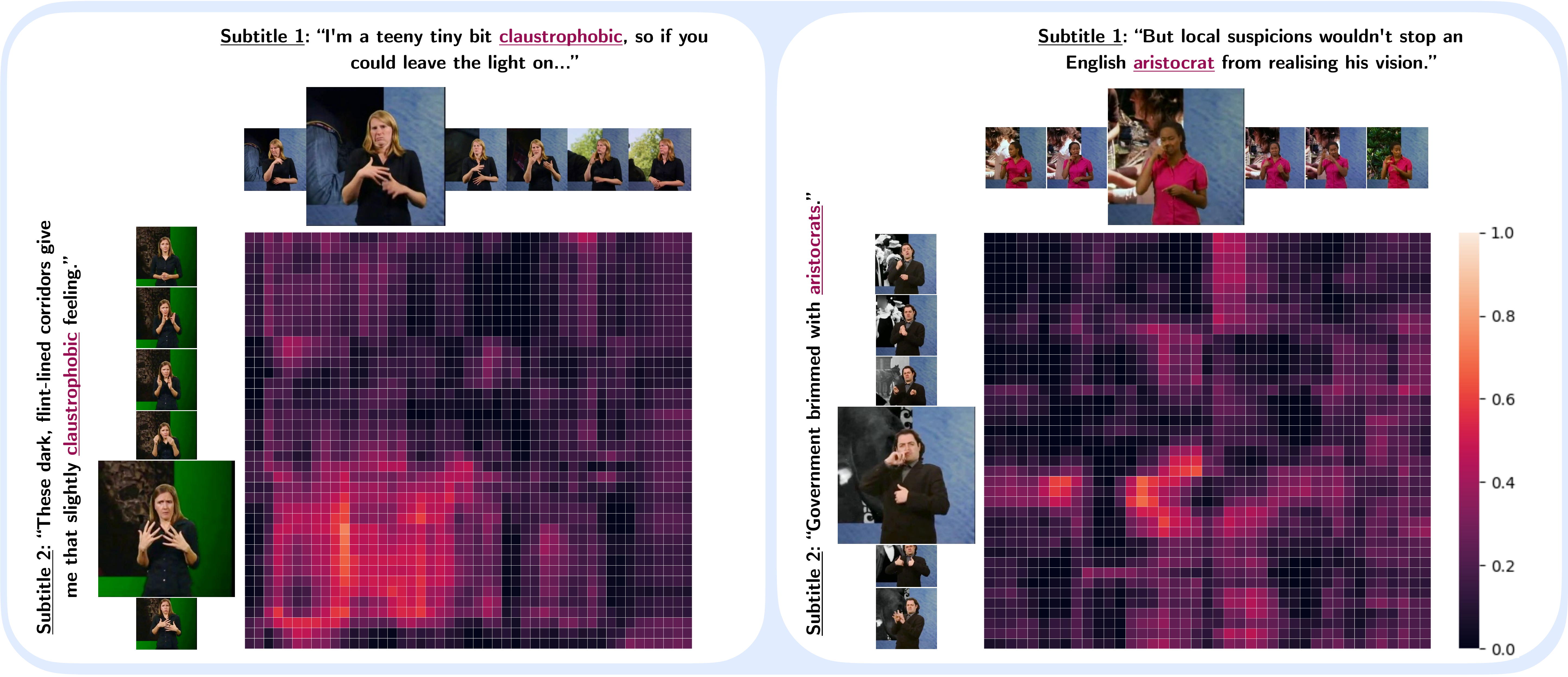}

    \caption{\textbf{Discovering novel sign classes (N)}: For two pairs of continuous signing sentences, we plot the score maps (as described in Sec.~\ref{subsec:exemplars}) between their feature sequences. We highlight the ability of our approach to spot novel sign classes.}

    \label{fig:cosine_similiarity}
\end{figure}

\begin{table}[t]
    \centering
    \caption{\textbf{Ablation on mining exemplar-based spottings for known signs \spotExemplarKnown{}:} We perform different ablations for mining known signs which have been unannotated by previous methods (\spotMnew{}, \spotDnew{}, A, \spotP{}). We experiment with the source of exemplar data (same episode, same signer, all data), the confidence of exemplar signs (0,0.5,0.8), the number of samples of exemplar data (5,10,20) and the pooling mechanism (average, max, vote). We evaluate on the test set (SENT-TEST).}
    \setlength{\tabcolsep}{4pt}
    \resizebox{\linewidth}{!}{
    \begin{tabular}{lllrl|ccc|ccc|ccc}
    \toprule
    &&&&& \multicolumn{3}{|c|}{\textbf{Training set}} &\multicolumn{3}{|c|}{\textbf{Spottings [full]}}& \multicolumn{3}{|c}{\textbf{MLP predictions [8K]}} \\
    Ann. & ex. & ex. & ex. & ex. & \multicolumn{1}{c}{full} & \#ann. & \#ann.&\multicolumn{3}{|c|}{\textbf{(subtitle dependent)}} & \multicolumn{3}{|c}{\textbf{(subtitle independent)}} \\
    src. & data & thres & \# & pooling & vocab & [full] & [8K] & Recall & IoU & Coverage & Recall & IoU & Coverage\\
    \midrule
    \spotExemplarKnown & same ep.    & 0  & var &  avg & 11.6K &  869K &  833K & 10.4 &  9.6 &  5.8 & 25.1 & 6.9 & 15.3 \\
    \spotExemplarKnown & same signer & 0  &  20 &  avg & 15.9K &  505K &  421K &  7.8 &  7.5 &  4.4 & 23.1 & 5.6 & 14.2 \\
    \spotExemplarKnown & all         & 0  &  20 &  avg & 16.7K &  351K &  252K &  5.7 &  5.7 &  3.3 & 21.5 & 5.1 & 13.4 \\
    \midrule
    \spotExemplarKnown & all         & 0.5 &  20 &  avg & 16.6K &  370K &  261K &  5.9 &  5.8 &  3.4 & 21.9 & 5.2 & 13.5 \\
    \spotExemplarKnown & all         & 0.8 &  20 &  avg & 16.6K &  458K &  358K &  7.4 &  7.3 &  4.3 & 25.2 & 6.2 & 15.7 \\
    \midrule
    \spotExemplarKnown & all         & 0.8 &  20 &  max & 15.4K & 1.48M & 1.38M & 20.2 & 18.6 & 10.8 & 27.6 & 8.4 & 17.7 \\
    \spotExemplarKnown & all         & 0.8 &  10 &  max & 15.4K & 1.07M &  982K & 15.2 & 14.0 &  8.3 & 27.9 & 8.0 & 17.7 \\
    \spotExemplarKnown & all         & 0.8 &   5 &  max & 15.3K &  740K &  664K & 10.7 & 10.0 &  6.0 & 27.6 & 7.6 & 17.4 \\
    \midrule
    \spotExemplarKnown & all      & 0.8 &  20 & vote & 15.9K      &   1.76M    &  1.63M    &  \textbf{25.8}   &   \textbf{23.3} &   \textbf{13.5} & \textbf{28.4}&\textbf{8.5} & \textbf{18.1}\\
    \spotExemplarKnown & all     & 0.8 &  10 & vote & 15.8K  &  1.32M     &   1.21M   & 20.0   &   18.1   &    10.7   & 28.4 & 8.3 & 18.1 \\
    \bottomrule
    \end{tabular}
    }
    
    \label{tab:exemplars}
\end{table}

\begin{table}[t]
    \centering
    \caption{\textbf{Pseudo-label spottings P \& Exemplar-based sign spottings for known \spotExemplarKnown{} and
        novel classes \spotExemplarNovel{}:} We highlight the boost in annotations by adding our pseudo-label annotations (P) as well as exemplar-based spottings of known (E) and novel (N) classes. We evaluate Spottings and MLP predictions on the test set (SENT-TEST). For the novel classes, we only show the evaluation of spottings since these are beyond the 8K training vocabulary of the MLP.}
    \setlength{\tabcolsep}{4pt}
    \resizebox{\linewidth}{!}{
    \begin{tabular}{l|ccc|ccc|ccc}
        \toprule
        & \multicolumn{3}{|c|}{\textbf{Training set}} &\multicolumn{3}{|c|}{\textbf{Spottings [full]}}& \multicolumn{3}{|c}{\textbf{MLP predictions [8K]}} \\
        & \multicolumn{1}{c}{full} & \#ann. & \#ann.&\multicolumn{3}{|c|}{\textbf{(subtitle dependent)}} & \multicolumn{3}{|c}{\textbf{(subtitle independent)}} \\
        Annotation source  & vocab & [full] & [8K] & Recall & IoU & Coverage & Recall & IoU & Coverage\\
        \midrule
        \spotMnew{} + \spotDnew{} + A~\cite{Varol21} + \spotP{}                                                                     & 20.9K & 3.64M & 3.56M & 25.4 & 23.5 & 13.9 & 29.8 & 8.9 & 19.2 \\
        \spotMnew{} + \spotDnew{} + A~\cite{Varol21} + \spotP{} + \spotExemplarKnown{}                                & 20.9K & 5.40M & 5.19M  & 45.3 & 40.7 & 23.3 & \textbf{30.7} & \textbf{9.5}& \textbf{19.8}\\
        \spotMnew{} + \spotDnew{} + A~\cite{Varol21} + \spotP{} + \spotExemplarKnown{} + \spotExemplarNovel{} & 24.8K & 5.47M &     - &\textbf{45.6} &\textbf{40.9} &\textbf{23.4}& - & - & - \\
        \bottomrule
        
    \end{tabular}
    }
    
    \label{tab:exemplars_with_novel}
\end{table}

\noindent{\textbf{Mining more Examples of Known and Novel Sign Classes with In-domain Exemplars.}} By explicitly querying words in the subtitle text which are not present in our annotations, we can obtain significantly more annotations. Tab.~\ref{tab:exemplars} shows multiple methods to use exemplar signs to find additional annotations for these signs. The best performing method takes spotting exemplars from across the whole training set, irrespective of signer or episode, and uses the voting scheme described in Sec.~\ref{subsec:exemplars} to localise signs. By using 20 spotting exemplars, we acquire 1.63M additional annotations. An MLP model trained \textit{only} on these additional annotations achieves a recall of 28.4 and coverage of 18.1. Tab.~\ref{tab:exemplars_with_novel} illustrates the impact of combining these additional annotations from spotting exemplars to \spotMnew{}, \spotDnew{}, A and \spotP{} annotations. With the additional exemplar-based annotations E, recall increases from 29.8 to 30.7 and coverage increases from 19.2 to 19.8, where the oracle recall and coverage are 86.7 and 55.2. Furthermore, by mining instances of novel sign classes N  (see Fig.~\ref{fig:cosine_similiarity}), we increase our total vocabulary to 24.8K and total number of annotations to 5.47M.

%% file: 06-conclusion.tex
\section{Conclusion}
\label{sec:conclusion}
Progress in sign language research has been accelerated in recent years due to the availability of large-scale datasets, in particular sourced from interpreted TV broadcasts. However, a major obstacle for the use of such data is the lack of available sign level annotations. Previous methods~\cite{Albanie20,Momeni20b,Varol21} only found \textit{sparse} correspondences between keywords in the subtitle and individual signs. In our work, we propose a framework which scales the number of confident automatic annotations from 670K to 5.47M (which we make publicly available). Potential future directions for research include: (1) increasing our number of annotations by incorporating context from \textit{surrounding} signing to resolve ambiguities; (2) investigating \textit{linguistic} differences between spoken English and British Sign Language such as the different word/sign ordering; (3) leveraging our automatic annotations for sign language translation.

%% file: 08-acknowledgements.tex
\noindent\textbf{Acknowledgements.} This work was supported by EPSRC grant ExTol, a Royal Society Research Professorship and the ANR project CorVis ANR-21-CE23-0003-01. LM would like to thank Sagar Vaze for helpful discussions. HB would like to thank Annelies Braffort and Michèle Gouiffès for the support.

%% file: 07-appendix.tex
\renewcommand{\thefigure}{A.\arabic{figure}} 
\setcounter{figure}{0} 
\renewcommand{\thetable}{A.\arabic{table}}
\setcounter{table}{0} 

\appendix

We first provide a discussion on the dependency and complementarity between different annotation approaches (Sec.~\ref{sec:comparison}). We subsequently describe implementation details in Sec.~\ref{sec:implementation}, present additional experimental results in Sec.~\ref{sec:experiments} and show some qualitative examples and failure cases in Sec.~\ref{sec:qualitative}. 

\section{Different automatic annotation approaches}\label{sec:comparison}

We provide a summary of the different approaches mentioned in this paper for annotating signs automatically in sign language interpreted TV shows, which consist of continuous signing and weakly-aligned English subtitles. We highlight specifically the limitations of different approaches and their dependencies. 
\begin{itemize}
\item M refers to automatic sign annotations obtained in previous work~\cite{Albanie20} from mouthings, as signers often mouth a word and sign it simultaneously. Specifically, the sign annotations are obtained by querying subtitle words in a signing window with a mouthing-based keyword spotting model and saving the most confident model predictions. Mouthing is a strong signal, but it cannot be used to annotate all data (since signers do not mouth continuously). Furthermore, these automatic annotations are skewed towards words with `easy' mouthings. 
\item D refers to automatic sign annotations obtained in previous work~\cite{Momeni20b} by leveraging online sign language dictionary clips. In more detail, a joint embedding space is learned between the \textit{isolated} dictionary video clips and the \textit{continuous} signing video sequences. At inference time, the cosine similarity between the continuous signing sequence and dictionary clips corresponding to subtitle words is calculated. The sign annotations correspond to the dictionary clips with highest similarity. Although these automatic annotations are not limited to signs accompanied by mouthings, they are limited to the vocabulary of the online sign dictionary. Furthermore, they are biased to an extent towards mouthings since the joint embedding space is learned using M annotations.
\item A refers to automatic sign annotations obtained in previous work~\cite{Varol21} by using the localisation ability from the attention mechanism of a video-to-text Transformer model. The encoder takes as input pre-computed video features (from a sign recognition model trained with M and D annotations)  and outputs a sequence of word stems. The sign annotations correspond to words which are correctly predicted and the sign timestamps are obtained by looking at the temporal position where the encoder-decoder attention is maximised. Compared to mouthing (M) and dictionary (D) annotations, the attention (A) annotations are obtained by taking context into account. 
\item \spotMnew{}  refers to new and improved mouthing annotations obtained in this work. In fact, we upgrade to a state-of-the-art keyword spotting model (Transpotter~\cite{prajwal2021visual}) and finetune this model on signer mouthings. We also use subtitles which are better aligned to the signing for centering our querying windows. This enables the number of detected mouthings and therefore automatic sign annotations to be greatly expanded.
\item \spotDnew{}  refers to new and improved dictionary annotations obtained in this work by (i) using subtitles which are better aligned to the signing for centering our querying windows, and (ii) expanding the query set to dictionary clips corresponding to similar words and synonyms to words in the subtitles. 
\item \spotP{} refers to new sign annotations obtained in this work through pseudo-labelling. In fact, we train a large-vocabulary (8K) sign classification model with automatic annotations from mouthings (M), dictionaries (D) and attention (A) and use it to pseudo-label. We firstly predict a sign class at each time step in a continuous signing video clip. We then filter the predicted signs to words in the corresponding subtitle.  

\item E and N are automatic sign annotations obtained in this work by relying on in-domain occurrences of signs. We localise a sign $w$ in a reference video $V_0$ given (i) the word corresponding to $w$ occurs in the subtitle associated with $V_0$, and (ii) other exemplar videos $V_1 \dots V_N$ with $w$ in the associated subtitles. When mining instances of \textit{known} classes E, the exemplar videos $V_1 \dots V_N$ are short video segments of the sign $w$ from previous annotation methods. When mining instances of \textit{novel} classes N, the exemplar videos $V_1 \dots V_N$ are longer, subtitle-length videos that have $w$ in their corresponding subtitle. E and N are collected by calculating a matrix of cosine similarities between video features of the reference and exemplar videos. These video features are extracted from the last layer of a sign recognition MLP model trained with \spotMnew{}, \spotDnew{}, A~\cite{Varol21}, and \spotP{} (see
\if\sepappendix1{Tab.~4}
\else{Tab.~\ref{tab:exemplars_with_novel}}
\fi
in main paper). These in-domain methods are necessary as not all signs have mouthing cues, and signs in continuous signing
may differ from their isolated realisations in dictionaries. 
\end{itemize}

\section{Implementation Details}\label{sec:implementation}

\subsection{Mining more Spottings through In-domain Exemplars (E)}

To mine for in-domain exemplars as described in
\if\sepappendix1{Sec.~3.1,}
\else{Sec.~\ref{subsec:exemplars},}
\fi
we choose $N$ video exemplars of spottings of signs that we wish to find in the reference video. For an exemplar sign, we choose 8 consecutive stride-4 features surrounding each spotting ($|\mathcal{C}_i|=8$ for $i=1\dots N$), where the features come from the last layer of the \spotMnew{} + \spotDnew{} + A~\cite{Varol21} + \spotP{} MLP model of
\if\sepappendix1{Tab.~5}
\else{Tab.~\ref{tab:exemplars_with_novel}}
\fi
of the main paper and are 256 dimensional. The values of $N$ are shown in the fourth column of
\if\sepappendix1{Tab.~3.}
\else{Tab.~\ref{tab:exemplars}.}
\fi
For the reference video, we choose a subtitle with 2s padding on either side, and use stride-4 features as candidate locations of signs. 

The methods `avg' and `max' noted in the fifth column of
\if\sepappendix1{Tab.~3}
\else{Tab.~\ref{tab:exemplars}}
\fi
are computed slightly differently to the method `vote' described in
\if\sepappendix1{Sec.~3.1.}
\else{Sec.~\ref{subsec:exemplars}.}
\fi
As before, we compute the cosine similarity between each feature at each position of the reference
video $c_0 \in \mathcal{C}_0$ and each position of the spottings exemplars $(c_1,\dots, c_n) \in \mathcal{C}_1\times \dots \times \mathcal{C}_N$. 
The cosine similarity is rescaled to the interval $[0,1]$. 
This results in $N$ score maps of dimension $|\mathcal{C}_0|\times |\mathcal{C}_i|$ for
$i=1\dots N$, which for us can be represented as a matrix
$\mathcal{M}$ of dimension $|\mathcal{C}_0|\times 8 \times N$. We take either the average or 
the maximum value of $\mathcal{M}$ over the $N$ exemplars to obtain a matrix $\mathcal{M}'$ of dimension $|\mathcal{C}_0|\times 8$. We then take the maximum of $|\mathcal{C}_0|\times 8$ across the exemplar temporal dimension to obtain a vector $L$ of dimension $|\mathcal{C}_0|$. We consider the first element of $L$ above a threshold $h$ to be the corresponding sign in the reference video. For the version where we take the average value of $\mathcal{M}$ over the $N$ exemplars, we let $h=0.7$; for the version where we take the maximum value of $\mathcal{M}$ over the $N$ exemplars, we let $h=0.8$. 

\subsection{Discovering Novel Sign Classes (N)}

In order to find a sign corresponding to a word $w$ in a reference video, we take $N=9$ positive exemplars corresponding to subtitles containing $w$, and $N'=27$ negative exemplars corresponding to subtitles not containing $w$. We do not use padding around either the reference video nor the exemplars. The confidences for these spottings correspond to the proportion of the $N$ exemplars with a cosine similarity match above a threshold $h$, i.e. the maximum value of $L^+$ as described in
\if\sepappendix1{Sec.~3.2.}
\else{Sec.~\ref{subsec:novel}.}
\fi
We consider all novel sign classes with a confidence threshold above 0, that is, with at least one match amongst the positive exemplars. 

\subsection{Synonym Collection}

We use synonyms both when querying keywords for spottings and when evaluating the performance of our MLP model. For these two purposes, we construct two different lists of synonyms. The first list is used for querying keywords for spottings and is large and flexible. The second list is a subset of the first; it is used to deem a prediction correct when evaluating our MLP model and is therefore more restrictive. 

The first list is an extensive list of synonyms combined from multiple sources: the online dictionaries SignBSL\footnote{\url{www.signbsl.com}}, and BSL~SignBank\footnote{\url{bslsignbank.ucl.ac.uk}} `related words' propositions for each sign video entry; words from the English synonym list from WordNet~\cite{wordnet} as well as words with GloVe~\cite{pennington-etal-2014-glove} cosine similarity above 0.9. In order to reduce noise, we remove synonyms with GloVe cosine similarity of less than 0.5. The second list of synonyms is a subset of the first, but we do not add all words with GloVe~\cite{pennington-etal-2014-glove} cosine similarity above 0.9. Instead, amongst words with GloVe similarity above 0.9, we keep only those predicted to be sign synonyms by a simple sign synonym detection model. The sign synonym model is a 4 layer MLP model predicting whether or not two video features correspond to the same or different signs. The model is trained on pairs of $M$+$D$+$A$ spottings from \cite{Albanie2021bobsl}, and evaluated using the validation split with 33 videos, rather than the 36 aligned test set episodes used in the rest of the paper. At evaluation, we search for sign synonyms from our first list only amongst words with GloVe similarity of 0.9 and above. For each potential pair or synonyms with more than 5 spottings in the evaluation set, we consider the pair to be sign synonyms if it is predicted to be identical for at least 50\% of the evaluation set examples. Tab.~\ref{tab:synonyms} shows examples of synonyms.

\begin{table}[]
\begin{tabular}{|l|l|}
\hline
\textbf{Word} & \textbf{Synonyms}                                                                                                                                                                                                                                                                          \\ \hline
change        & evolution, diversity, conversion, switch, variety, convert, other, acquire, \\
& transform, amend, transformation, deepen, selection, evolve, adaptation, \\
& alteration, amendment, various, adapt, transfer, become, exchange, alter, \\ & modify, variation, modification, vary, among, shift \\\hline
bus           & coach, heap, metro, subway, tube, underground, vehicle, bus stop                                                                                                                                                                                                                      \\\hline
rare          & uncommon, few                                                                                                                                                                                                                                                                              \\\hline
content       & message, capacity, substance, subject, context, insert, relief                                                                                                                                                                                                                             \\\hline
architect     & designer                                                                                                                                                                                                                                                                                   \\\hline
airplane      & aeroplane                                                                                                                                                                                                                                                                                  \\\hline
skyscraper    & city                                                                                                                                                                                                                                                                                       \\\hline
king          & royal, prince, princess, mogul, queen, power, tycoon, baron                                                                                                                          \\ \hline                                                                                               
\end{tabular}
    \caption{
    \textbf{Examples of synonyms}: Our list of synonyms contains English words with similar meaning or words that can be signed using the same sign. 
    \label{tab:synonyms}
    }
\end{table}

\subsection{Transpotter Finetuning}
In
\if\sepappendix1{Section~3.4}
\else{Section~\ref{subsec:improvements}}
\fi
of the main paper, we discuss the domain gap between the lip movements in videos with the audio track removed (for example, from TV programmes) and the mouthings in sign language videos. As the Transpotter~\cite{prajwal2021visual} is trained on the former, we finetune it on the pseudo-annotated sign language mouthings to reduce the domain gap. In this section, we describe the process of extracting pseudo annotations and the subsequent finetuning. 

\subsubsection{Extracting Pseudo-annotations:}
We start with a pre-defined list of keywords that are at least $3$ phonemes in length according to the CMU dictionary~\cite{cmu} and find all occurrences of these keywords in the subtitles. We take the video segment corresponding to the subtitle as our search window. We add 10 second padding (as also done in~\cite{Albanie2021bobsl}) on either side of this video segment to account for the temporal misalignment between the continuous signing and audio-aligned subtitles. We query for the keywords present in the subtitle in order to obtain the temporal localization of each keyword in the video segment. As the video segment is much longer than the segments seen by the model during training, we perform a windowed inference with short $3$ second windows. We have a $1.2$ second overlap between successive windows. We run two windowed passes through the video, where the start time of the second pass is delayed by one second. This is to ensure that in at least one of these passes, the desired sign (often $< 1$ second in length) occurs completely within the short window. The Transpotter outputs a per-frame probability indicating whether a word is uttered at that frame. We save the frame number with the maximum probability as a possible annotation for the word and later filter these annotations based on confidence values.

\subsubsection{Finetuning:} As described in the main paper, we perform two rounds of finetuning. We first extract pseudo-labels using the Transpotter model from~\cite{prajwal2021visual}, pretrained on silent speech videos. We filter the mouthings with a confidence $\ge 0.7$ as positive samples. In each batch, we oversample negative word-video pairs, in order to reduce false positives. We finetune the pre-trained Transpotter at a low learning rate of $1e^{-6}$ using the AdamW optimizer~\cite{loshchilov2017decoupled}. After convergence, we extract annotations with this more accurate finetuned model. We finetune the model a second time using the same hyper-parameters as above but resuming from the model weights from the first stage of finetuning. Further rounds of finetuning bring negligible improvements. Our final mouthing annotations $M^*$ are extracted using this model.

\subsubsection{How Does Finetuning Help?} 
\begin{figure}[t]
    \centering
    \noindent\includegraphics[width=\textwidth]{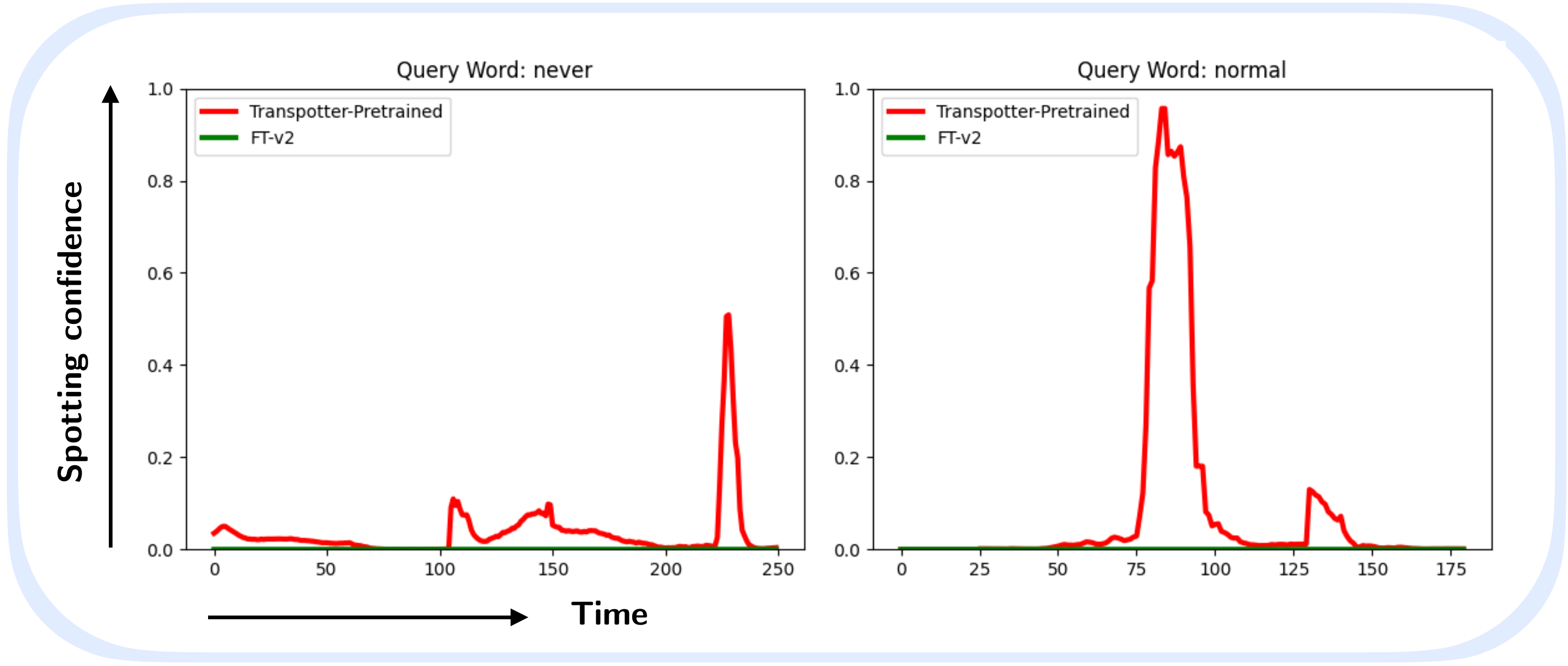}
    \caption{\textbf{Finetuning the Transpotter on pseudo-annotations leads to fewer false positive detections}: We show two qualitative examples to illustrate the impact of finetuning the mouthing model. For a given query word and a short video segment, we plot the per-frame confidence scores of the two models, i.e. before and after finetuning. We can see that the pre-trained Transpotter spots mouthings even though they are not present, whereas the finetuned model correctly predicts near-zero confidence, indicating that the word is indeed not mouthed in the given video segments.}
    \label{fig:mouthing_conf}
\end{figure}

We observe that the Transpotter pre-trained on silent speech segments produces a large number of false detections on signing video segments as shown in Fig~\ref{fig:mouthing_conf}. 

After finetuning, the model is less likely to erroneously predict a query word. The decreased number of false positives is reflected by a reduction in overall size of the automatically annotated dataset, noted in
\if\sepappendix1{Tab.~2}
\else{Tab.~\ref{tab:improved-mouthing-dict-pl}}
\fi
of the main paper. The finetuned model only spots $412K$ mouthings compared to the pre-trained model's $661K$. Despite a $1.5\times$ reduction in dataset size, the MLP model achieves better performance when trained on the $412K$ mouthings. Thus, finetuning the Transpotter improves downstream task performance, while also enabling faster and more efficient training of our MLP classifier due to fewer training samples.

\subsection{Video Backbone (I3D)}
\label{app:subsec:i3d}
Here, we describe the training of our I3D video backbone, which is used as the frozen feature extractor and as the source of pseudo-labelling for sign spotting. 

As shown in Tab.~\ref{tab:features}, we start with the M+D baseline from~\cite{Albanie2021bobsl} which is initialised with Kinetics~\cite{kinetics} pretraining. This model is trained with sign annotations with confidence above 0.8, resulting in 426K training samples from a 2,281 sized vocabulary. The model takes as input 16 consecutive video frames at 25 fps and a cropped $224 \times 224$ spatial region (from an initial $256 \times 256$ region). The input to the model is therefore $3 \times 16 \times 224 \times 224$, since our frames are RGB. For each sign annotation from mouthing (M), a sequence of 16 contiguous frames is randomly sampled from a window covering 15 frames before the time associated with the annotation and 4 frames after the annotation, i.e.,
$[-15, 4]$ around the mouthing peak. For dictionary annotations,
the window around the similarity peak is $[-3, 22]$. I3D is trained for 25 epochs using SGD with
momentum (with a momentum value of 0.9), with a batch-size of
size 4. An initial learning rate of 0.01 is decayed by a factor of
10 after 20 epochs. Augmentations are applied during training including spatial cropping and color augmentations as well as scale and horizontal flip augmentations. The model produces a 1024-dimensional embedding (following average pooling) which is passed to our last linear layer, which outputs scores with the dimensionality of the number of classes. When evaluating the I3D predictions, I3D is run in a sliding window manner over the continuous signing with a stride of 4. 

We explore how changing our pretraining effects performance: instead of only pretraining on Kinetics, we use a publicly released model (available on the webpage for~\cite{Varol21}) which is first pretrained on Kinetics then finetuned on BSL1K~\cite{Albanie20} on a 5K vocabulary size. As shown in Tab.~\ref{tab:features}, this marginally improves performance on our downstream task of continuous sign recognition.

We explore how expanding the vocabulary from 2K to 8K varies performance: this increases the number of training instances with confidence over 0.8 from 426K to 670K. In this case, our model is only trained for 17 epochs (due to computational costs) with an initial learning rate of 3e-2, reduced by a factor of 10 at epoch 12. As shown in Tab.~\ref{tab:features}, this increases our recall from 25.5 to 26.3 and coverage from 15.5 to 16.3. This final model is chosen as our frozen feature extractor and as our source of pseudo-labelling for sign spotting: both features and class predictions are obtained by running I3D in a sliding window fashion with a stride of 4.

\begin{table}[t]
    \centering
    \setlength{\tabcolsep}{25pt}
    \caption{
    \textbf{Video backbone (I3D)}: We highlight the improved performance of I3D on the test set (SENT-TEST) when trained on a larger vocabulary (8K instead of 2K) and better pretraining using BSL1K~\cite{Albanie20,Varol21}. 
    }
    \resizebox{\linewidth}{!}{
    \begin{tabular}{l|c|c|c|c|c|c}
        \toprule
         &  &  & &\multicolumn{3}{|c}{\textbf{I3D predictions}} \\
          &  & &  &\multicolumn{3}{|c}{\textbf{(subtitle independent)}} \\
        Annot. source & Pre-training& Num. I3D train annot. & Vocab. size & Recall & IoU & Coverage \\
        \midrule
        M~\cite{Albanie20}+D~\cite{Momeni20b}  \cite{Albanie2021bobsl} &Kinetics~\cite{kinetics} & 426K & 2K & 25.3  & 6.3 & 15.4\\
       M~\cite{Albanie20}+D~\cite{Momeni20b}&Kinetics~\cite{kinetics}+BSL1K~\cite{Albanie20,Varol21} & 426K & 2K & 25.5 & 6.4 & 15.5\\
         M~\cite{Albanie20}+D~\cite{Momeni20b}&Kinetics~\cite{kinetics}+BSL1K~\cite{Albanie20,Varol21}   & 670K & 8K&\textbf{26.3} & \textbf{7.9}&\textbf{16.3} \\
        \bottomrule
    \end{tabular}
     \label{tab:features}
    }
    
\end{table}

\subsection{Lightweight Classifier (MLP)}
\label{app:subsec:mlp}
As new sets of spottings are generated, a light weight MLP classifier is trained on the pre-extracted I3D features. Our 4-layer MLP module has layers of dimension (1024,512,256,8K) where the last layer corresponds to the number of sign classes and contains LeakyRelu activations in between. The first linear layer also has a residual connection on the 1024-dimensional I3D input features. The MLP is trained with a batch size of 128 for 15 epochs, with the learning rate initially set to 1e-2 and decayed by a factor of 10 at epochs 5 and 10. When evaluating the MLP predictions, the MLP is run in a sliding window fashion, outputting one feature for each I3D input feature (where the I3D features are extracted with a stride of 4).

\section{Additional Experiments}\label{sec:experiments}

\subsection{Varying Spotting Confidence}
We show how varying the spotting confidence impacts the quality of spottings from previous methods, versus the improved spottings proposed in our work. As shown in Tab.~\ref{tab:confidence}, even when reducing the confidence, our improved \spotMnew{} + \spotDnew{} give the best performance on our downstream task.
\begin{table}[t]
    \setlength{\tabcolsep}{4pt}
     \centering
    \caption{\textbf{Varying spotting confidence:} We highlight how varying the spotting confidence changes both our Spotting and MLP prediction evaluation performance. We evaluate on the test set (SENT-TEST).}
    \resizebox{\linewidth}{!}{
    \begin{tabular}{l|ccc|ccc|ccc}
        \toprule
        & \multicolumn{3}{|c|}{\textbf{Training set}} &\multicolumn{3}{|c|}{\textbf{Spottings [full]}}& \multicolumn{3}{|c}{\textbf{MLP predictions [8K]}} \\
        & \multicolumn{1}{c}{full} & \#ann. & \#ann.&\multicolumn{3}{|c|}{\textbf{(subtitle dependent)}} & \multicolumn{3}{|c}{\textbf{(subtitle independent)}} \\
        Annotation source  & vocab & [full] & [8K] & Recall & IoU & Coverage & Recall & IoU & Coverage\\
        \midrule
        M(0.8) + D(0.8) & 15.0K & 680K & 670K  &8.5 & 8.3& 4.8& 24.9 & 7.1 & 15.5\\
        M(0.8) + D(0.75) & 15.9K & 1.90M & 1.76M & 18.4 & 16.9& 9.8 & 23.3 & 8.1 & 15.1\\
        M(0.8) + D(0.7) & 16.6K & 5.22M & 4.86M & 33.1 & 29.2& 16.6&18.9 & 7.8 & 13.4\\
        M(0.5) + D(0.7) & 24.7K & 5.74M & 5.32M  & 35.3 & 30.9& 17.5 &19.1 & 7.8 & 13.4\\
        M(0.5) + D(0.7) + A(0) & 24.7K & 6.17M & 5.74M & 37.0 & 32.5 & 18.3 & 20.3 & 8.3 & 13.9 \\
      \spotMnew{}(0.8) + \spotDnew{}(0.8) &  20.9K & 2.00M & 1.94M & 19.0 & 17.6 & 10.5 & 29.0 & 7.9 & 18.4 \\
       \spotMnew{}(0.8) + \spotDnew{}(0.75) & 21.7K & 7.89M & 7.77M  & 41.9& 36.7& 24.0&27.0 & 7.7 & 18.5 \\
        \spotMnew{}(0.5) + \spotDnew{}(0.75) & 22.4K & 7.97M & 7.84M & 42.4 & 37.1& 24.3&27.2 & 7.8 & 18.6 \\
        \spotMnew{}(0.5) + \spotDnew{}(0.75) + A(0) & 22.5K & 8.40M & 8.28M  &43.8 &38.3 & 24.8&27.5 & 7.9 & 18.7 \\
        \bottomrule
    \end{tabular}
    }
    \label{tab:confidence}
\end{table}

\section{Qualitative examples}\label{sec:qualitative}

\subsection{Densification Visualisations}
\begin{figure}[t]
\centering
\noindent\includegraphics[width=\textwidth]{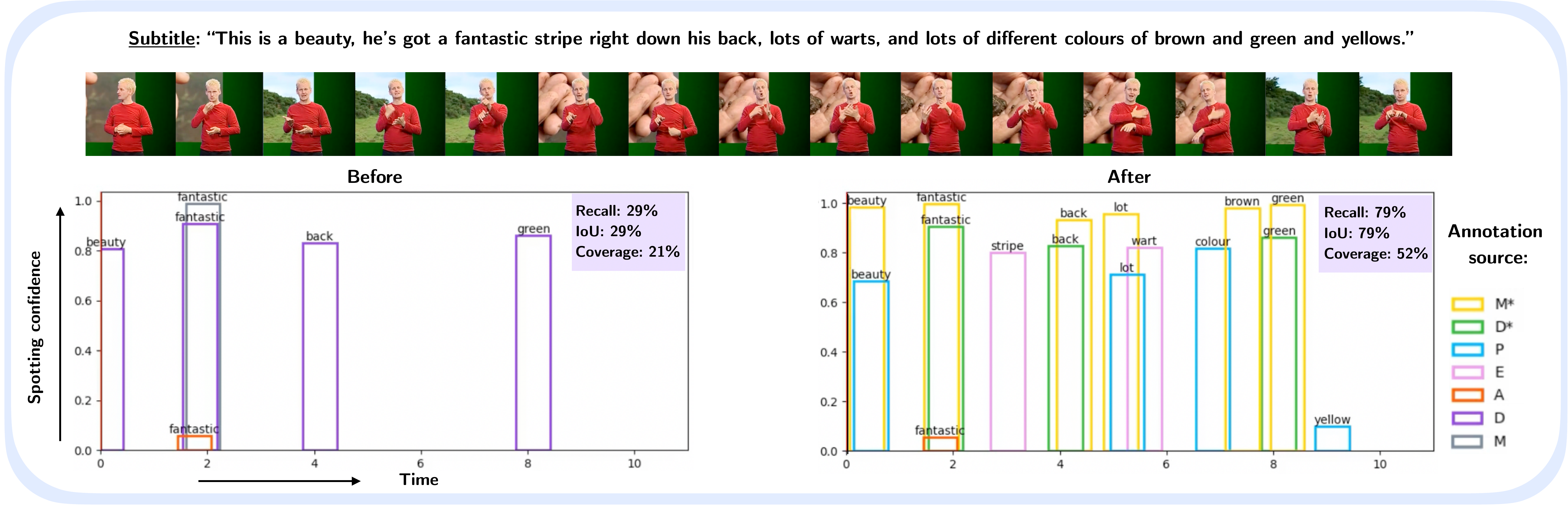}
\noindent\includegraphics[width=\textwidth]{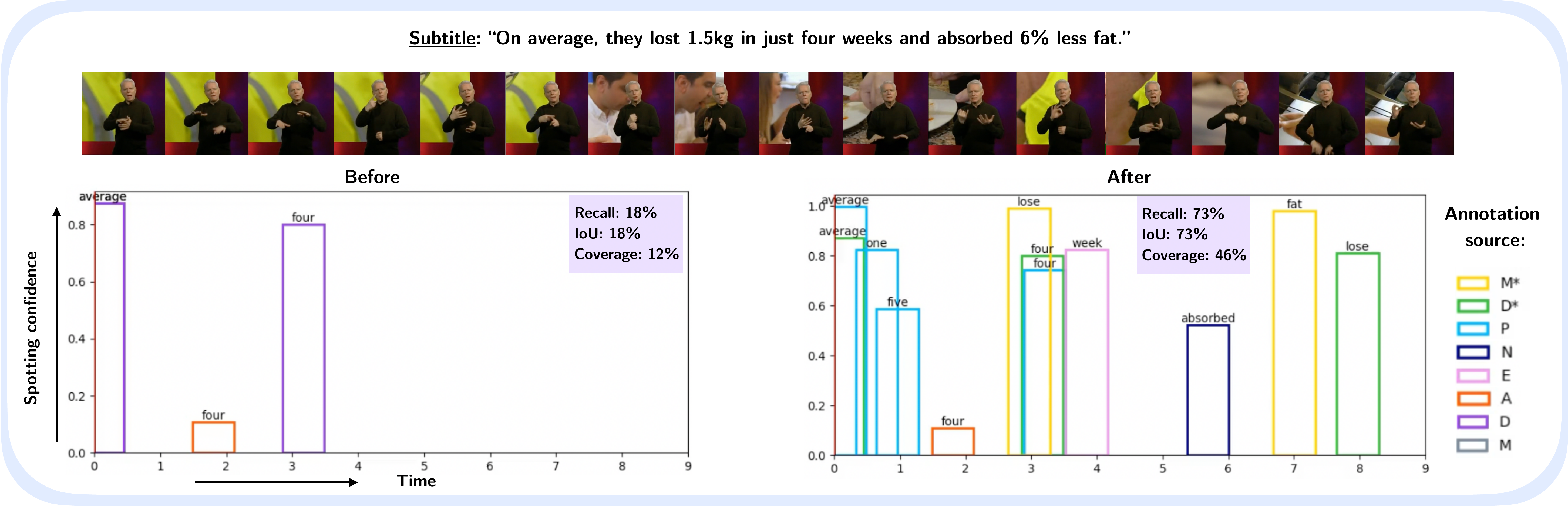}
 \caption{\textbf{Densification}: For two continuous signing sequences, we show plots of automatic sign annotation timelines, along with their confidence and annotation source, \textit{before} and \textit{after} our framework is applied. We observe that our method enables \textit{densification} by two measures: removing gaps in the timeline so that we have a dense signing sequence spotted; and also increasing the number of words in the corresponding spoken language subtitle we recall. M, D, A refer to spottings obtained from previous methods from mouthings~\cite{Albanie20}, dictionaries~\cite{Momeni20b} and attentions~\cite{Varol21} respectively. $M^{*}$, $D^{*}$, P,  E, N refer to new and improved spottings from mouthings~\cite{prajwal2021visual}, dictionaries~\cite{Momeni20b}, I3D sign recognition pseudo-labels, in-domain exemplar spottings of known sign classes as well as in-domain exemplar spottings of novel classes respectively.}
    \label{fig:densify}
\end{figure}

In Fig.~\ref{fig:densify}, we show visualisations of our densified sign sequences after our framework is applied.

\subsection{Known Classes Spottings Visualisations}
\begin{figure}[t]
    \centering
    \noindent\includegraphics[width=\textwidth]{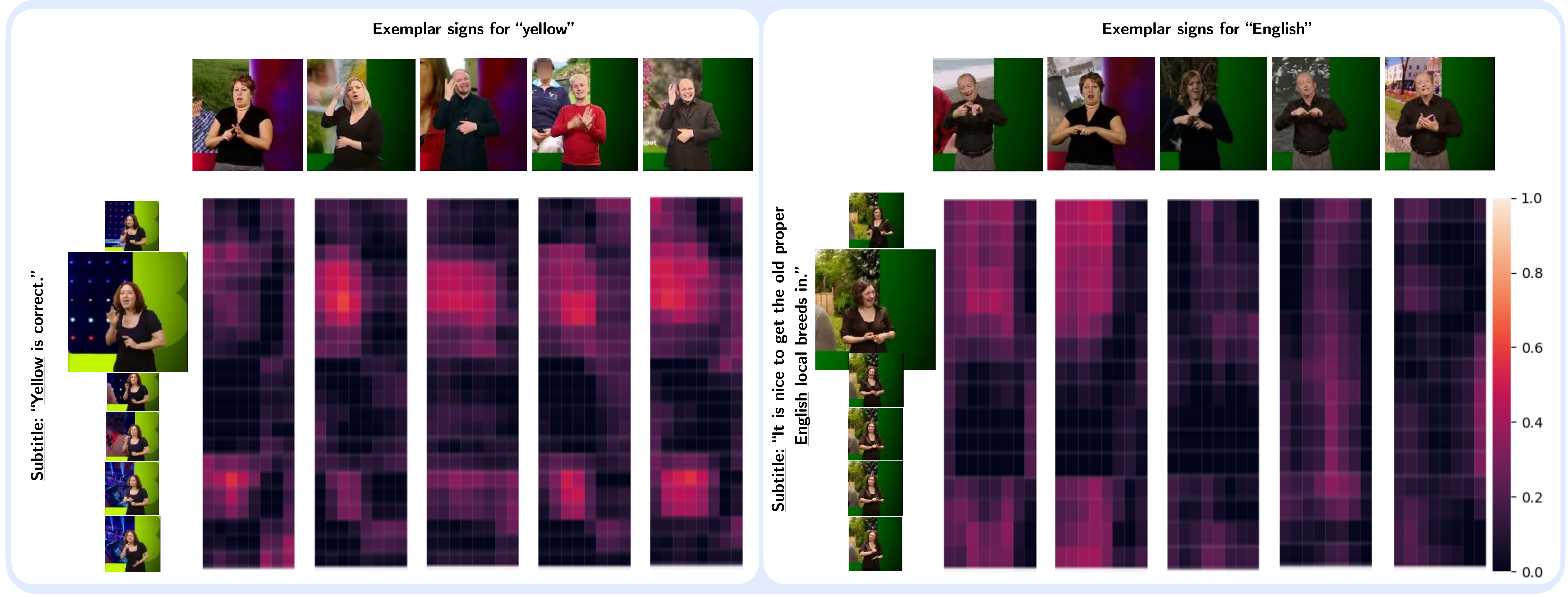}
    \caption{\textbf{Mining with spotting exemplars}: By comparing the score maps between a subtitle text and multiple spotting exemplars, we can temporally locate a lexical sign in a video segment. The left example illustrates how we can find the sign for `yellow'. There are two different signs for `yellow', where the second, third and fifth exemplars correspond to the sign used in the subtitle, and the first and fourth exemplars show an alternative sign. By using a voting method, we can count the number of exemplars with a high cosine similarity at a particular temporal location in the reference subtitle. The right example searches for the sign `English' in a subtitle using 5 exemplars. The fifth exemplar in an incorrect spotting annotation, and has a low cosine similarity. However, with enough exemplars by different signers in different contexts, we can locate likely temporal locations of a sign in a subtitle.}
    \label{fig:known}
\end{figure}

In Fig.~\ref{fig:known}, we show visualisations of our score maps for annotating instances of known classes through our in-domain exemplar signs. 

\subsection{Novel Classes Spottings Visualisations}
\begin{figure}[t]
    \centering
    \noindent\includegraphics[width=\textwidth]{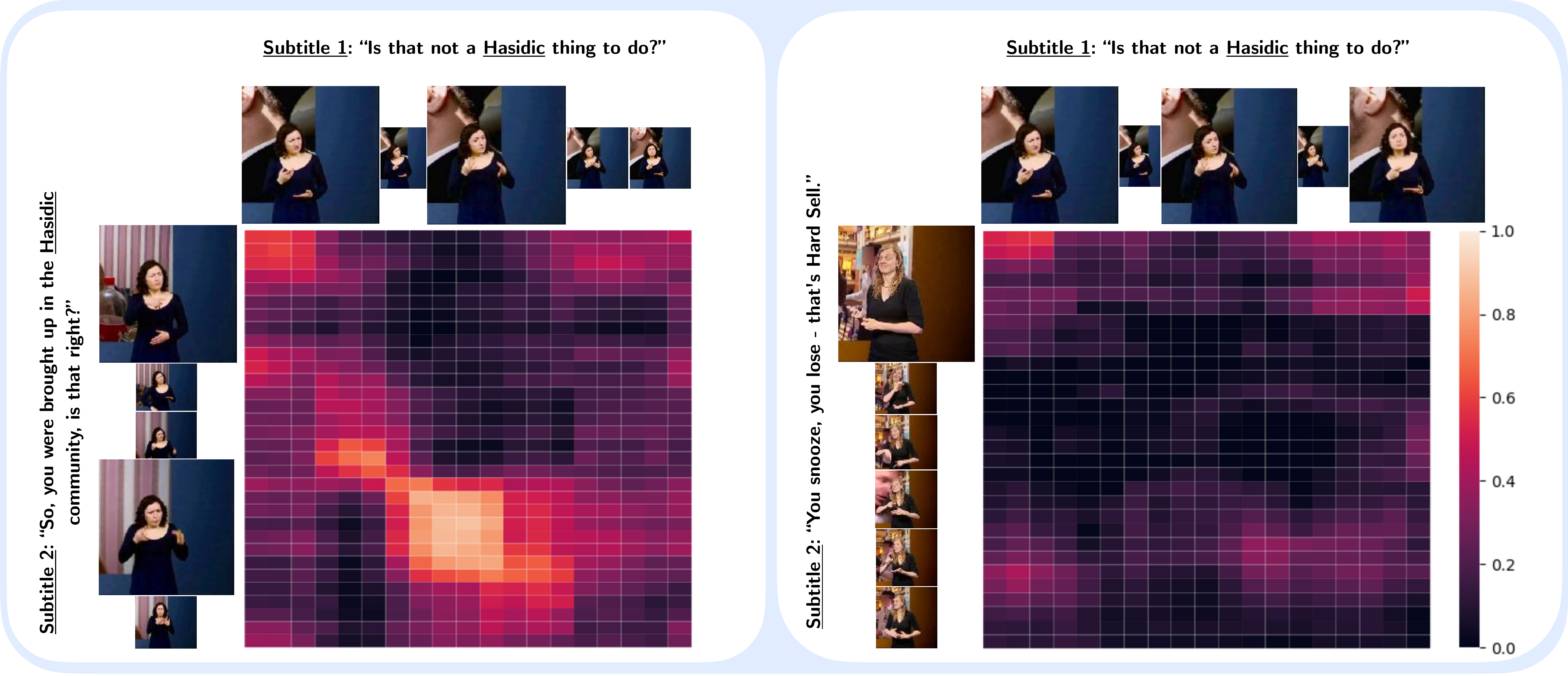}
    \caption{\textbf{Necessity of negative samples}: On the left, we show the cosine similarity between features of two subtitles, both of which contain the word `Hasidic'. The cosine similarity is indeed high at the temporal intersection of both signs for `Hasidic'; but the cosine similarity does also peak at pointing signs common to both subtitles. On the right, we show a score map for a subtitle containing the word `Hasidic' and a subtitle without this keyword. By using the score maps of negative examples, we can identify non-lexical signs common across subtitles, such as pointing signs, and hence avoid incorrectly labeling the common lexical query sign. 
    }
    \label{fig:novel1}
\end{figure}

\begin{figure}[t]
    \centering
    \noindent\includegraphics[width=\textwidth]{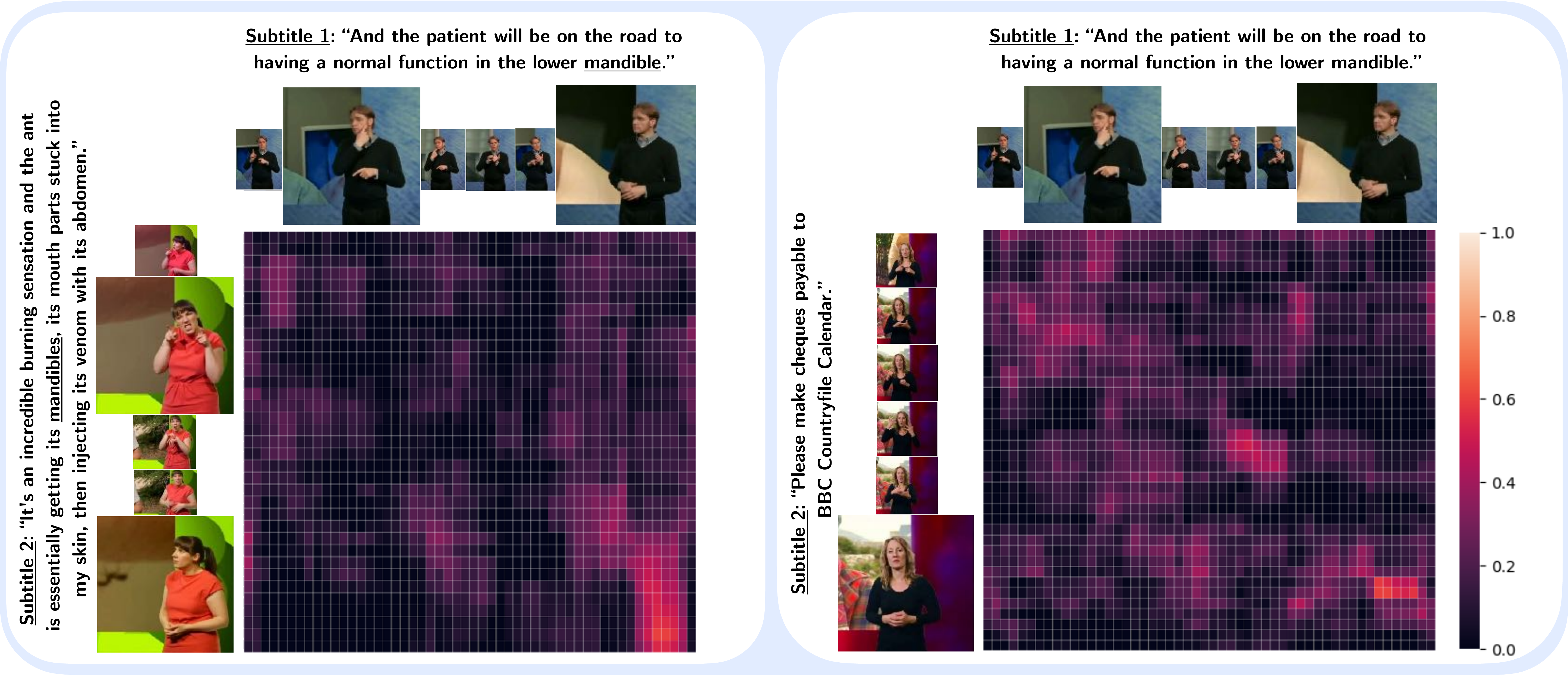}    
    \caption{\textbf{Failure case}: On the left, we show the score map for two subtitles sharing the common word `mandible'. However, in the first example, `mandible' refers to a human mandible and in the second example, mandibles of an ant. The sign language interpretation of this word differs in each context, and the score map only shows strong cosine similarity when the signers are in a neutral pause position. The right score map demonstrates that this neutral position, frequent across many subtitles, can be located using negative exemplars. Using information from negative exemplars, we can avoid incorrect annotations. 
    }
    \label{fig:novel2}
\end{figure}

We show visualisations of score maps for annotating instances of novel classes through our in-domain weak exemplar subtitles. 
Fig.~\ref{fig:novel1} illustrates the necessity of using negative samples to avoid incorrectly identifying signs common to many subtitles such as pointing signs, pause gestures or other common gestures as the common lexical sign across exemplars. Fig.~\ref{fig:novel2} shows a failure case, where we cannot identify the sign for `mandible' due to two different realisations of the sign depending on context.

